\pdfoutput=1 
\RequirePackage{fix-cm}
\documentclass[twocolumn]{svjour3}          
\smartqed  
\usepackage{ifpdf}
\usepackage{cite}
\usepackage[cmex10]{amsmath}
\usepackage{algorithmic}
\usepackage{array}
\usepackage{fixltx2e}
\usepackage{url}
\usepackage{graphicx}
\usepackage{float}
\usepackage{amsmath}
\usepackage{amssymb}

\usepackage{color}
\usepackage{ulem}
\usepackage{ifthen}
\usepackage{natbib}
\usepackage{hyperref}
\hypersetup{
  colorlinks=true,
  citecolor=blue,
  linkcolor=red,
  urlcolor=blue
}

\newcommand{\ie}{i.e.,\ }   
\newcommand{\figname}{Fig.}
\newcommand{\tabname}{Table}
\newcommand{\secname}{Section\ }

\newcommand{\bw}{\mathbf{w}}
\newcommand{\bx}{\mathbf{x}}

\newcommand{\bff}{\mathbf{f}}

\newcommand{\bbR}{{\mathbb{R}}}

\newcommand{\TP}{n\textsc{tp}}
\newcommand{\FP}{n\textsc{fp}}
\newcommand{\FN}{n\textsc{fn}}
\newcommand{\preci}{\textsc{precision}}
\newcommand{\reca}{\textsc{recall}}

\newcommand{\sym}{SYM}

\newcommand{\Rrwm}{Object}

\newcommand{\vgg}{VGG}
\newcommand{\snu}{Co-reg}

\definecolor{gray}{rgb}{0.5,0.5,0.5}
\definecolor{green}{rgb}{0, 0.6, 0}
\definecolor{orange}{rgb}{1, 0.5, 0} 	
\definecolor{mahogany}{rgb}{0.75, 0.25, 0.0}
\definecolor{purple}{rgb}{0.6, 0, 0.6}
\definecolor{darkgreen}{rgb}{0, 0.4, 0}

\newboolean{revising}
\setboolean{revising}{false}

\ifthenelse{\boolean{revising}} {

\newcommand{\delete}[1]{\textcolor{red}{\sout{#1}}}

\newcommand{\comment}[1]{\textcolor{red}{[{\bf comment:} {#1}]} }
}{

\newcommand{\delete}[1]{}

\newcommand{\comment}[1]{}
}

\usepackage{epstopdf}
\begin{document}

\title{Descriptor Ensemble: An Unsupervised Approach\\ to Descriptor Fusion in the Homography Space
}
\subtitle{}


\author{Yuan-Ting Hu         \and
        Yen-Yu Lin		     \and
        Hsin-Yi Chen         \and
        Kuang-Jui Hsu        \and
        Bing-Yu Chen
}


\institute{Yuan-Ting Hu \and Yen-Yu Lin \and Kuang-Jui Hsu \at
           Research Center for Information Technology Innovation, Academia Sinica, Taipei, Taiwan\\
              \email{r01922042@ntu.edu.tw, yylin@citi.sinica.edu.tw, and kjhsu@citi.sinica.edu.tw.}
           \and
           Hsin-Yi Chen \and Bing-Yu Chen \at
           Department of Computer Science and Information Engineering, National Taiwan University, Taipei, Taiwan\\
           \email{fensi@cmlab.csie.ntu.edu.tw and robin@ntu.edu.tw.}
}


\maketitle

\begin{abstract}
With the aim to improve the performance of feature matching, we present an unsupervised approach to fuse various local descriptors in the space of homographies. Inspired by the observation that the homographies of correct feature correspondences vary smoothly along the spatial domain, our approach stands on the unsupervised nature of feature matching, and can select a good descriptor for matching each feature point. Specifically, the {\em homography space} serves as the common domain, in which a correspondence obtained by any descriptor is considered as a point, for integrating various heterogeneous descriptors. Both geometric coherence and spatial continuity among correspondences are considered via computing their {\em geodesic distances} in the space. In this way, mutual verification across different descriptors is allowed, and correct correspondences will be highlighted with a high degree of consistency (\ie short geodesic distances here). It follows that {\em one-class SVM} can be applied to identifying these correct correspondences, and boosts the performance of feature matching. The proposed approach is comprehensively compared with the state-of-the-art approaches, and evaluated on four benchmarks of image matching. The promising results manifest its effectiveness.
\end{abstract}




\section{Introduction}

Image matching aims to identify common regions across images. As a key component of image content analysis, image matching has attracted great attention for several years. It is one of the critical stages in widespread image processing and computer vision applications, such as panoramic stitching~\citep{Szeliski97}, object recognition~\citep{Lowe04}, image retrieval~\citep{Datta08}, and common pattern discovery~\citep{Liu10}.

\begin{figure}[tH]
\begin{center}
\hspace*{-0.15 cm}
\begin{tabular}{cc}
\includegraphics[height = 0.60 in]{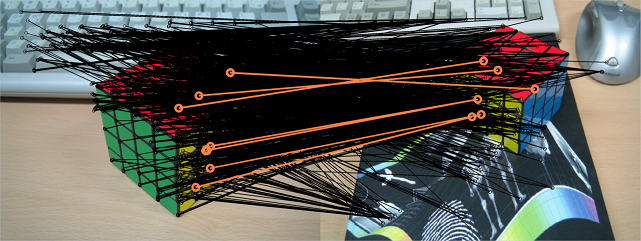}&\hspace{-0.3 cm}\includegraphics[height = 0.60 in]{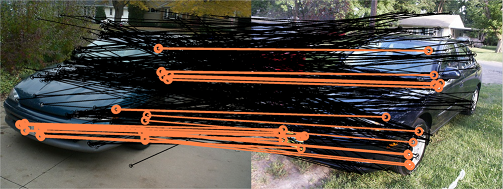}\\[5pt]
(a) SIFT & \hspace{-0.3 cm}(e) SIFT \\[5pt]
\includegraphics[height = 0.60 in]{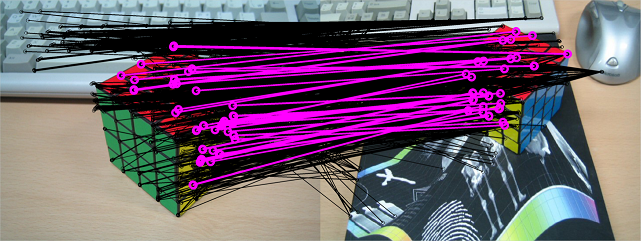}&\hspace{-0.3 cm}\includegraphics[height = 0.60 in]{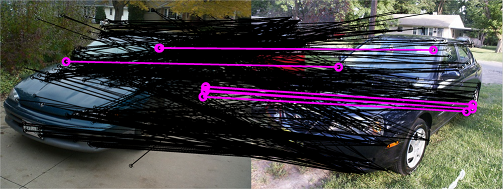}\\[5pt]
(b) Raw intensities & \hspace{-0.3 cm}(f) Raw intensities\\[5pt]
\includegraphics[height = 0.60 in]{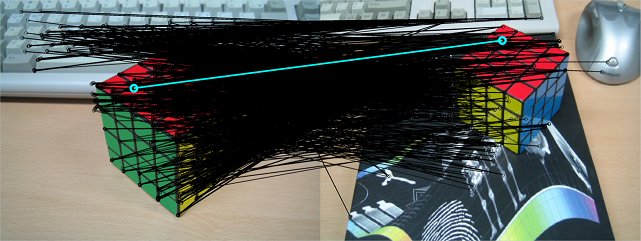} & \hspace{-0.3 cm}\includegraphics[height = 0.60 in]{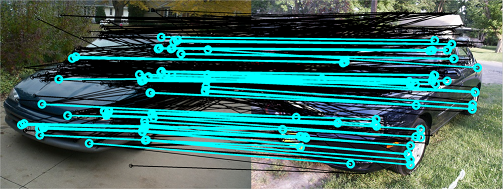}\\[5pt]
(c) Geometric blur & \hspace{-0.3 cm}(g) Geometric blur\\[5pt]
\includegraphics[height = 0.60 in]{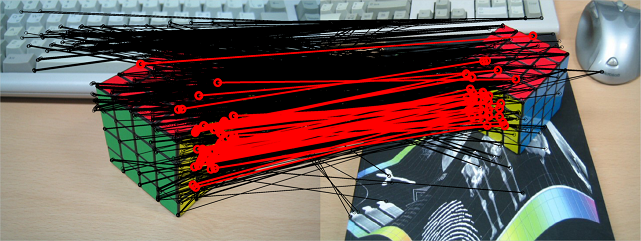} & \hspace{-0.3 cm}\includegraphics[height = 0.60 in]{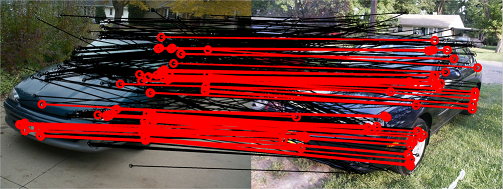}\\[5pt]
(d) Ours & \hspace{-0.3 cm}(h) Ours\\[0pt]
\end{tabular}
\end{center}
\caption{Feature matching on two image pairs, (a) $\sim$ (d) {\tt magic cube} and (e) $\sim$ (h) {\tt car}. The matching results by using three different descriptors, including SIFT, raw intensities, and geometric blur, are shown in the first three rows, respectively. While correct correspondences are drawn in a specific color, wrong ones are in black. In {\tt magic cube}, color/intensity information is important for matching owing to the high degree of color coherence. In contrast, shape and gradient features are more reliable in {\tt car}. This example indicates that the performance of a descriptor varies from image to image. In addition, the deficiency of using a single descriptor is revealed. Our approach instead makes use of multiple, complementary descriptors, and can achieve superior matching results, as shown in (d) and (h).}
\label{fig:tradeoff}
\end{figure}

\begin{figure*}[tH]
\begin{center}
\begin{tabular}{ccc}
	\hspace*{-0.6 cm}
	\begin{tabular}{c}
		\begin{tabular}{cc}
			\includegraphics[height = 0.85 in]{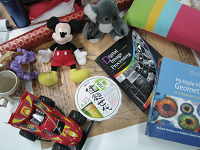}& \hspace*{-0.45 cm}
			\includegraphics[height = 0.85 in]{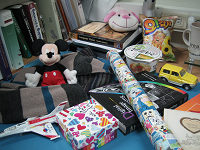}\\
			(a)&(b)\\[5pt]
		\end{tabular}\\[5pt]
			\includegraphics[height = 0.85 in]{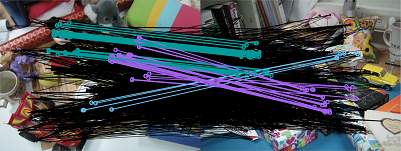}\\
			(c) SIFT\\
	\end{tabular}&
	\hspace*{-1.05 cm}
	\begin{tabular}{c}
		\includegraphics[height = 0.85 in]{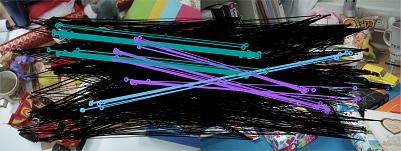}\\
		(d) Raw intensities\\[5pt]
		\includegraphics[height = 0.85 in]{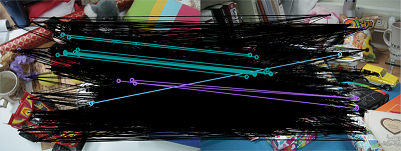}\\[0pt]
		(e) Geometric blur\\
	\end{tabular}& \hspace*{-0.9 cm}
	\begin{tabular}{c}
		\vspace*{+0.5 cm}\fbox{\includegraphics[height = 1.855 in, width = 2.00 in ]{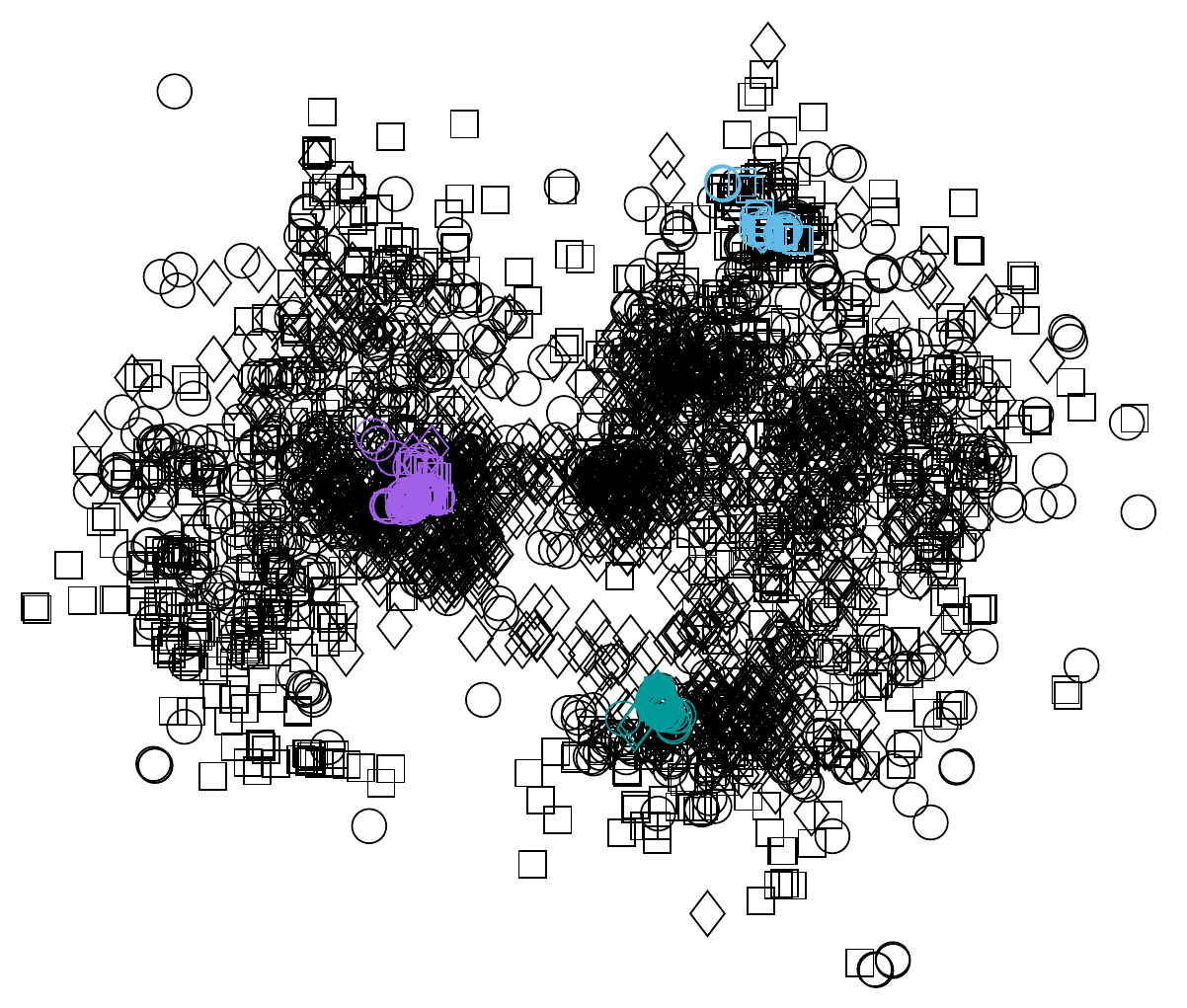}}\\
		[-10pt](f)\\
	\end{tabular}
\end{tabular}
\end{center}
\caption{(a) $\&$ (b) Two input images for feature matching. (c) Matching results by using SIFT. (d) Matching results by using raw intensities. (e) Matching results by using geometric blur. Each wrong correspondence is drawn in black, while each correct one is in a specifical color depending on the common object that it resides in. (f) The $2$D visualization of correspondences in the homography space via multi-dimensional scaling (MDS). The circle, square, and diamond markers denote correspondences obtained by SIFT, raw intensities, and geometric blur, respectively. This figure demonstrates  that not only geometric coherence but also spatial continuity are highly relevant the correctness of correspondences.}
\label{fig:overview}
\end{figure*}

Coupling interest points with local feature descriptors has been proven to be an effective way of image matching~\citep{Mikolajczyk05a,Mikolajczyk05b}. Although the development of powerful descriptors~\citep[e.g.,][]{Lowe04,Berg01,Belongie02,Bay06,Tola10,Wang11,Hauagge12} has gained significant progress, there is in general no a single descriptor that is sufficient for dealing with all kinds of variations and deformations in feature matching. Most descriptors are designed on the trade-off between {\em distinctiveness} and {\em invariance}. The more distinctive the descriptor is, the higher precision but the lower recall it may get. On the contrary, descriptors with high degrees of invariance often result in high recall but low precision. It implies that the goodness of a descriptor is usually {\em image-dependent}. Without any prior knowledge about images, using only one descriptor becomes insufficient and unreliable to conquer the wild image matching problems.

\figname~\ref{fig:tradeoff} shows the matching results on two image pairs, {\tt magic cube} and {\tt car}, by using three descriptors, {\em SIFT}~\citep{Lowe04}, {\em raw intensities}, and {\em geometric blur}~\citep{Berg01}, and our approach, respectively. The strong color coherence presents in the case of {\tt magic cube}, so the color-based descriptor, raw intensities, gives good results. On the other hand, better performance is achieved by the shape-based descriptors, SIFT and geometric blur, in the case of {\tt car}. However, none of the three descriptors perform well in both the two cases. This example points out not only the performance fluctuation of a descriptor among images but also the deficiency of using a single descriptor.

In view of these issues, we aim at improving the quality of image matching with the aid of multiple, complementary descriptors. Two challenges arise in this scenario. First, features extracted by different descriptors are usually of different dimensions and with different scales of statistics. Even their adopted metrics for similarity measure are diverse. How to effectively fuse heterogeneous descriptors becomes a challenging problem. Second, image matching in general is an unsupervised task. The goodness of descriptors is hard to evaluate without ground truth. When feature matchings by different descriptors present, how to identify correct ones from them is another problem. In this paper, we present an unsupervised approach that can effectively overcome the two problems, and generate accurate and dense correspondences by leveraging complementary descriptors.

The idea of our approach is illustrated through \figname~\ref{fig:overview}. \figname~\ref{fig:overview}(a) and \ref{fig:overview}(b) show two input images for matching. The matching results by three descriptors, including SIFT, raw intensities, and geometric blur, are plotted in \figname~\ref{fig:overview}(c), \ref{fig:overview}(d), and \ref{fig:overview}(e), respectively. Wrong correspondences are drawn in black. Each correct correspondence is displayed with a specific color according to the common object that it resides in. Despite the varieties, a correspondence by any descriptor can be specified by its {\em geometric transformation} (or {\em homography}) in the same way. It implies that correspondences by all descriptors can be treated as points in the homography space, which can then serve as the common domain for fusing heterogeneous descriptors. We compute the pair-wise distances among points (correspondences), and show them in \figname~\ref{fig:overview}(f) via {\em multi-dimensional scaling} (MDS)~\citep{Cox94}, which summarizes high-dimensional data in a low-dimensional space by taking the pair-wise distances as input. The circle, square, and diamond markers denote correspondences obtained by using SIFT, raw intensities, and geometric blur, respectively. It can be observed in \figname~\ref{fig:overview}(f) that correct (colored) correspondences on the same object share similar homographies no matter by which descriptors they are established. They hence gather together in the homography space, while incorrect correspondences distribute irregularly. This observation suggests that geometric consistency among correspondences is highly relevant to their correctness. Moreover, \figname~\ref{fig:overview}(f) also indicates that correct correspondences are spatially correlated, since only correct correspondences within the same objects are geometrically consistent. The details of the adopted homography space and the similar measure between correspondences will be introduced later.

Inspired by the observation in \figname\ref{fig:overview}, this work carries out feature matching with multiple descriptors, and can distinguish itself with the following three main contributions. First, we propose to use the {\em homography space} as a unified domain, for fusing photometric and geometric information caught by descriptors, where correspondences obtained by all descriptors are represented as points. In this way, the unified representation of correspondences enables geometric checking across diverse descriptors. To our knowledge, this is novel in the field. Second, we present an approach that stands on the unsupervised nature of image matching. It can determine the correctness of correspondences, and choose an appropriate descriptor for matching each feature point without any training data. Specifically, we estimate the geometric and spatial consistency among correspondences via computing {\em geodesic distances}~\citep{Tenenbaum00} on a designed graph to smoothly transfer the carried information in the homography space. Through this process, correct correspondences are highlighted with strong coherence with each other. It follows that {\em one-class SVM}~\citep{Scholkopf01} can be applied to picking these correct correspondences. Third, our approach is comprehensively evaluated on four benchmarks of image matching, and jointly takes five descriptors into account, including {\em SIFT}~\citep{Lowe04}, {\em LIOP}~\citep{Wang11}, {\em DAISY}~\citep{Tola10}, {\em geometric blur} (GB)~\citep{Berg01} and {\em raw intensities} (RI). Our approach is compared with four image matching baselines and four baselines of descriptor fusion. It achieves significantly better results than those by the best descriptors and baselines in most cases. We hence term our approach {\em Descriptor Ensemble} in the sense it combines multiple, complementary descriptors, and improves the performance of image matching.

The rest of this manuscript is organized as follows. A review of the related works is given in {\secname}\ref{sec:relatedWork}. The problem we address is stated in {\secname}\ref{sec:PD}. We described the adopted homograph space and the used similarity measure between correspondences in {\secname}\ref{sec:Homo}. The proposed approach is specified in {\secname}\ref{sec:our}. The experimental setup and results are given in {\secname}\ref{sec:imp} and {\secname}\ref{sec:exp}, respectively. Finally, we conclude this work in {\secname}\ref{sec:con}.

\section{Related Work\label{sec:relatedWork}}

The literature on image feature matching is quite extensive. Our review focuses on those crucial to the development of the proposed approach.

\subsection{Local Feature Descriptors}

Local feature descriptors~\citep{Mikolajczyk05b} have been extensively studied, especially since the seminal works by \citet{Schmid97} and \citet{Lowe04}. Various descriptors have been designed to be robust to noises while invariant to particular types of deformations in matching. For example, {\em SIFT (scale-invariant feature transform)}~\citep{Lowe04} describes image regions in the gradient domain, constructing an $128$-dimensional histogram, and is known to be robust to scale and orientation changes. {\em LIOP (local intensity order pattern)}~\citep{Wang11} encodes both the local and global ordinal information, and can alleviate the unfavorable variations caused by the changes of lighting conditions. {\em DAISY}~\citep{Tola10} is featured with fast feature extraction, while keeps invariant to viewpoint changes. In addition, diverse visual cues have been explored in descriptor construction, such as color characteristics~\citep{Abdel-Hakim06}, shapes~\citep{Berg01}, internal self-similarities~\citep{Shechtman07}, topological information~\citep{Lobaton11}, and local symmetries~\citep{Hauagge12}. These descriptors are designed on the trade-off between distinctiveness and invariance. Thus, there does not exist an optimal descriptor in a wide range of test images. By contrast, we introduce our approach into image matching by employing multiple descriptors to complement one another and, thus solve this problem.

Instead of using handcrafted descriptors, a branch of research efforts has been made to automatically derive discriminative descriptors by machine learning techniques~\citep[e.g.,][]{Varma07,Hua07,Winder07,Carneiro10}. \citet{Hua07} introduced discriminant learning into constructing feature descriptors by minimizing the distances of correct matching pairs and maximizing the distances of non-matched ones. \citet{Varma07} found an optimal tradeoff for a given classification problem and proposed a kernel learning method of learning transformations of features from training image pairs. \citet{Carneiro10} proposed a {\em universal feature transform} that maps patches into robust descriptors. The universal feature transform can be incrementally updated, and hence expand its applicability. However, additional training data are required in these approaches. Moreover, since the optimal descriptors for matching vary from image to image, the performances of these learned descriptors highly rely on the consistency between training images and test images. Instead, our approach provides a fully unsupervised way to select proper descriptors. Neither additional training data nor prior knowledge about the images are needed.

\subsection{Correspondence Verification}

Identifying correct feature correspondences from candidates is an important step in image matching. Geometric layout checking is one of the most effective ways, because the geometric layout of feature correspondences often reveals their correctness. {\em RANSAC}~\citep{Fischler81} is a classic method for removing outliers through geometric verification. It estimates a global transformation and rejects outliers simultaneously. A correspondence is considered as an outlier and deleted if it is inconsistent with the transformation that the majority agree. One advantage of RANSAC is its easy implementation to fulfill geometric verification. However, RANSAC is not able to deal with multiple object matching and non-rigid transformations, and would be computationally expensive when the number of outliers becomes large.

\citet{Chui03,Ma14} and~\citet{Pang14} relaxed the geometric assumption of correspondences from obeying a global transformation to a smooth feature mapping function. The feature points are linked to their corresponding points through a smooth mapping function, which makes a non-rigid transformation expressible.~\citet{Ma14} and~\citet{Pang14} have demonstrated an effective way to determine the parameter values of the mapping function with the {\em vector field}. They can identify the correct correspondences. However, owing to the smoothness assumption of the mapping function, these methods are not very robust in matching multiple objects, especially when the transformations of multiple objects vary greatly.

Instead of deriving the transformations involved in matching, non-rigid deformations can be dealt with via measuring the similarities between correspondences, since correct correspondences tend to be consistent with each other. Both the photometric information given by descriptors and the geometric relationship of correspondences can be used in similarity computation. Some examples of similarity measures can be found in works by~\citet{Leordeanu05,Zheng06,Choi09} and~\citet{Liu10}. With the similarities between correspondences, a branch of research efforts~\citep{Zheng06,Cour06,Choi09,Cho10,Leordeanu12,Li13} cast the task of correct correspondence identification as an optimization problem over the matching score. {\em Graph matching}~\citep{Cour06,Cho10,Leordeanu12} is one of the representative techniques in optimizing the matching score. Although it is an NP hard problem, various graph matching algorithms have been proposed to get the approximate solutions. Nonetheless, these approaches are sensitive to outliers, and are less robust in multiple object matching.

Clustering based techniques for grouping correspondences with geometric constraints have been explored. \citet{Cho09} established a linkage model of correspondence clusters, and iteratively merged the clusters based on their geometric consistency. \citet{Zhang12} refined and reformulated the linkage model as a directed graph to further eliminate ambiguousness. The computational efficiency might be an issue due to the iterative algorithms. \citet{Liu10} found local maximizers on the matching score and merged them if any two of them are similar enough. The clustering framework can handle multiple transformations, but the parameter values of the developed models are difficult to determine, such as the thresholds of identifying the correct clusters, the criteria of merging clusters and the scale factor in~\citep{Liu10}.

Voting schemes can be used for measuring the consistency between correspondences. The correspondences are checked by the pair-wise geometric consistency via mutual voting among correspondences.~\citet{Avrithis14} transformed correspondences into Hough space and the voting results are collected efficiently with a pyramid structure.~\citet{Chen13} cast the voting process as a kernel density estimation problem in the transformation space. These approaches can identify multiple objects effectively. However, voting methods would become less powerful to find correct correspondences when the number of correct matchings is so less that votes from them are not dominant during voting. Our proposed approach in spirit follows the idea of project correspondences into transformation space, but we use the geodesic distance to include the spatial layout for improving the geometric consistency measurement. The most important feature of our approach is that multiple descriptors are considered. The proposed approach increases the number of correct matchings, and can avoid the situation mentioned above.

\subsection{Multiple Descriptor Fusion}

Since different descriptors can catch diverse visual cues, using multiple descriptors has been a feasible way for improving performance. A number of approaches~\citep[such as][]{Bosch07,Mortensen05,Zhang07,Harada10,Brox11,Weinzaepfel13} have been developed to fuse diverse descriptors for improving image matching, classification and alignment. \citet{Mortensen05} proposed to {\em concatenate} SIFT~\citep{Lowe04} and shape context~\citep{Belongie02}, and reported good results in matching. However, simple feature concatenation ignores the possible variations of feature dimensions and feature value scales among descriptors. It may lead to suboptimal performance, especially when less powerful descriptors have dominant feature dimensions or values. To address this problem, \citet{Bosch07} represented images under each descriptor as a {\em kernel matrix}. The works by \citet{Brox11} and \citet{Weinzaepfel13} integrated descriptor matching for handling large displacement in optical flow, and combined multiple visual evidences represented in form of {\em energy functions}. Although kernel matrices and energy functions can serve as the unified domains for descriptor fusion, these approaches tune or learn fixed weights for descriptor combination. It may not be suitable for image matching, because the optimal descriptors change from image to image. Furthermore, using brute force search to determine descriptor weights may become infeasible, when there are a large number of descriptors to be considered. Besides, image matching is an unsupervised task, and no training or validation data are available for determining descriptor weights in general.

Our approach tackles these issues, and has the following two advantages: 1) Multiple descriptors are represented by their homographies in matching so that we can work with complementary descriptors without worrying about their diversities of feature dimensions or feature value scales; 2) Our approach allows geometric checking across descriptors, and consensus correspondences will reveal through the process. It means that the plausible correspondences by various descriptors can be jointly identified in a fully unsupervised manner.

\section{Problem Statement\label{sec:PD}}

We aim to match two given images $I^P$ and $I^Q$, which come with the sets of detected feature points, $U^P = \{u_{i}^P\}_{i=1}^{N^P}$ and $U^Q = \{u_{i}^Q\}_{i=1}^{N^Q}$, respectively. The support region of each feature $u_i \in U^P \cup U^Q$ is assumed to an ellipse in this work. These feature points can be obtained by using off-the-shelf detectors, such as {\em Harris-Affine}~\citep{Mikolajczyk04}, {\em Hessian-Affine}~\citep{Mikolajczyk04}, the salient region detector~\citep{Kadir04}, or their combinations. We use Hessian-Affine detector for its efficiency and high repeatability. Multiple descriptors are employed to characterize each feature point. The center and the described appearances of feature $u_i$ are respectively denoted by $\bx_i$ and $\{\bff_{i,m}\}_{m=1}^M$, where $M$ is the number of the employed descriptors. The set $\tilde{{\cal C}} = U^P \times U^Q $ covers all possible feature {\em correspondences} (or {\em matchings}). Our goal is to detect correct correspondences in $\tilde{{\cal C}}$ as many as possible.

The number of the detected feature points in a image, \ie $N^P$ or $N^Q$, is often in the order of $10^3$. Directly searching correct correspondences in $\tilde{{\cal C}}$ may be inefficient. Hence we start from compiling a reduced set ${\cal C}$ of $\tilde{{\cal C}}$ by filtering out correspondences that are unlikely to be correct. For each feature $u_i^{P} \in I^P$, we find the set of the most similar $r$ matchings, ${\cal C}_{i,m} = \{(u_i^{P}, u^Q_{i_{k}} \in I^Q)\}^r_{k=1}$, with descriptor $m$ and the yielded distance $\|\bff_{i,m}^P - \bff_{i_k,m}^Q\|$. Since total $M$ descriptors are adopted, at most $r \times M$ matchings of $u_i^{P}$ are kept in ${\cal C}$ after removing the duplicates. Namely,
\begin{equation}
{\cal C} = \bigcup_{i=1}^{N^P} {\cal C}_{i}, \mbox{ where } {\cal C}_{i} = \bigcup_{m=1}^{M} {\cal C}_{i,m}. \label{eq:C}
\end{equation}
We will work on the reduced candidate set ${\cal C}$. The value of $r$ controls the trade-off between efficiency and accuracy. It is empirically set as $5$ in all our experiments. 

\section{Homography Space: A Common Domain for Descriptor Fusion\label{sec:Homo}}

In this section, we first introduce how to compute the geometric transformations of  correspondences and how to measure their geometric dissimilarity. Then, we show how to fuse diverse descriptors in the homography space, where correspondences obtained by different descriptors are treated in a unified manner.


\subsection{Homographies of Feature Correspondences}

The elliptical region of feature $u_{i}$ can be specified by mapping a circular region centered on the origin via the affine transformation:
\begin{equation}
T(u_{i}) = \left[
\begin{array}{cc}
A(u_{i})    &  \bx_{i} \\
\mathbf{0}^{\top}   &  1
\end{array}
\right] \in \bbR^{3 \times 3},
\label{eq:nor_pat}
\end{equation}
where $\bx_i \in \bbR^{2 \times 1}$ is the feature center, and $A(u_{i}) \in \bbR^{2 \times 2}$ is a non-singular matrix which accounts for the scale, the shape, and the orientation of $u_{i}$. After normalization with transformation $T(u_{i})^{-1}$, all the adopted descriptors can be applied to $u_{i}$, and generate $\{\bff_{i,m}\}_{m=1}^M$. Refer to work by~\citet{Mikolajczyk04} for the details.



A homography in this work refers to the geometric transformation of a feature correspondence. For a correspondence between $u^P_{i} \in U^P$ and $u^Q_{j} \in U^Q$, the transformation from the support region of $u^P_{i}$ to that of $u^Q_{j}$ can be derived as
\begin{equation}
H_{ij} = T(u^Q_{j})*T(u^P_{i})^{-1} \in \bbR^{3 \times 3}.
\label{eq:rel_tra}
\end{equation}
$H_{ij}$ is a $6$-dof (degrees of freedom) affine transformation. Thus, it can be viewed as a point in the $6$-dimensional homography space ${\cal H}$.

Consider two correspondences $c = (u^P_{i}, u^Q_{j}) \in {\cal C}$ and $c' = (u^P_{i'}, u^Q_{j'}) \in {\cal C}$. We use the {\em reprojection error} to measure their geometric dissimilarity. Specifically, the homography matrices, $H_{ij}$ and $H_{i'j'}$, of the two correspondences are firstly computed by Eq. (\ref{eq:rel_tra}). The {\em projection error} of $(u^P_{i}, u^Q_{j})$ with respect to $H_{i'j'}$ is then calculated by
\begin{equation}
d_{err}(u^P_{i}, u^Q_{j}, H_{i'j'}) = \| \bx_j^Q - \rho(H_{i'j'}  \left[  \begin{array}{c} \bx_i^P \\ 1 \end{array} \right])\|,
\end{equation}
where function $\rho: \bbR^3 \rightarrow \bbR^2$ is defined as
\begin{equation}
\rho( \left[\begin{array}{c} a \\ b \\ c \end{array}\right]) = \left[ \begin{array}{c} \frac{a}{c} \vspace{0.08in}\\ \frac{b}{c} \end{array} \right].
\end{equation}
The projection error $d_{err}(u^P_{i}, u^Q_{j}, H_{i'j'})$ is the induced error when changing the homography from $H_{ij}$ to $H_{i'j'}$ on correspondence $(u^P_{i}, u^Q_{j})$. The reprojection error between correspondences $c$ and $c'$ is then defined as
\begin{align}
d(c, c') = &\frac{1}{4} \big( d_{err}(u^P_{i}, u^Q_{j}, H_{i'j'}) + d_{err}(u^Q_{j}, u^P_{i}, H_{i'j'}^{-1}) \nonumber\\
&\hspace{2pt}+ d_{err}(u^P_{i'}, u^Q_{j'}, H_{ij}) + d_{err}(u^Q_{j'}, u^P_{i'}, H_{ij}^{-1}) \big). \label{eq:reproj_err}
\end{align}
We will use the reprojection error to measure the geometric dissimilarity between correspondences in ${\cal C}$.

\subsection{Homography as A Unified Representation}

In view of the complexity of feature matching, we consider characterizing each feature point $u_i$ with total $M$ kinds of different descriptors, \ie $\{\bff_{i,m} \in {\cal F}_m\}_{m=1}^M$, and each descriptor is associated with a measure of photometric dissimilarity $d_m: {\cal F}_m \times {\cal F}_m \rightarrow \bbR$.


The resulting representations by these descriptors are typically of various dimensions and even in diverse forms, such as vectors, histograms, and pyramids. The associated dissimilarity measures may be also different, such as Euclidean distance, $\chi^2$ distance or negative cosine similarity. To avoid the difficulties caused by these varieties, a unified representation of these descriptors is preferred.

It can be observed that the homography of a feature correspondence in Eq. (\ref{eq:rel_tra}) is {\em descriptor-independent}, and can hence serve as the domain for descriptor fusion. That is, each descriptor determines its own candidate correspondences as shown in Eq. (\ref{eq:C}), while all the candidate correspondences are represented by the corresponding homographies, each of which can be treated as a point in the homography space ${\cal H}$ of six dimensions. In this way, the dissimilarity between correspondences that are generated by different descriptors can be measured through Eq. (\ref{eq:reproj_err}). It implies that knowledge, such as the geometric distribution of matchings, can be transferred across descriptors. The proposed approach will leverage this nice property to work with complementary descriptors, and boost the performance of image matching.

\section{The Proposed Approach\label{sec:our}}

We formulate the task of image matching as finding an appropriate matching for each feature point $u_i^P$ in image $I^P$, \ie picking the most plausible correspondence from ${\cal C}_i$ in Eq. (\ref{eq:C}). In this work, multiple descriptors collaborate in the sense that they jointly determine candidate correspondences with diverse region characteristics and different kinds of invariance, so the probability that the correct correspondence resides in ${\cal C}_i$ largely increases. The goal at this stage is to determine the correct correspondence for each $u_i^P$, if it exists. The unsupervised nature of image matching makes this task very challenging, because no prior knowledge or relevant training data are available.


We tackle this issue based on the observation that the homographies of correct correspondences vary smoothly along the spatial locations in the image. Specifically, we firstly employ a graph to encode the spatial arrangement among correspondences, and compute the {\em geodesic distances}~\citep{Tenenbaum00} on the graph for geometric coherence estimation. Then, we utilize {\em one-class SVM}~\citep{Scholkopf01} to identify correct correspondences, since it can effectively separates alike (both geometrically and spatially coherent here) data from the outliers. The two steps are respectively described in the following.

\subsection{Geometric and Spatial Coherence Estimation\label{sec:our-geo}}

To jointly consider the geometric and spatial relationships among correspondences, we construct a graph ${\cal G} = ({\cal V}, {\cal E})$, in which each correspondence $c_i \in {\cal C}$ is associated with a vertex $v_i \in {\cal V}$, while an undirected edge $e_{ij} \in {\cal E}$ is added to connect vertices $v_i$ and $v_j$ if the end points in image $I^P$ of $c_i$ and $c_j$ are near enough. That is, the number of vertices in ${\cal G}$, $|{\cal V}|$, is the same as the number of correspondences in ${\cal C}$. In our implementation, we consider two feature points in $I^P$ are nearby if one point belongs to the $k$ spatial nearest neighbors of the other point. Weight $w_{ij}$ assigned to edge $e_{ij}$ is defined as
\begin{equation}
w_{ij} =
    \begin{cases}
    d(c_i, c_j),& \text{if $e_{ij} \in {\cal E}$,}   \\%
    \infty,& \text{otherwise,}
    \end{cases}
    \label{eqn:eij_weight}
\end{equation}
where the geometric dissimilarity $d(c_i, c_j)$ between correspondences $c_i$ and $c_j$ is given in Eq. (\ref{eq:reproj_err}). With the weighted graph, we compute the geodesic distance between each pair of vertices (\ie correspondences). We denote the geodesic distance between correspondences $c_i$ and $c_j$ by $d_{geo}(c_i,c_j)$ hereafter. It can be seen that graph ${\cal G}$ integrates the spatial continuity into the estimation of geometric coherence. The use of geodesic distance catches the phenomenon that the homographies of correct correspondences on the same object may change smoothly along their spatial locations. Therefore, the resulting dissimilarity measure can deal with possible deformations in matching.

Suppose there are $N$ correspondence candidates, \ie ${\cal C} = \{c_1, ..., c_N\}$. We compute the pair-wise geodesic distances $\{d_{geo}(c_i,c_j)\}_{i,j=1}^N$. The correct correspondences will be highlighted with strong geometric and spatial coherence (short geodesic distances) with other correct correspondences. It is worth pointing out that compared with $d(c_i, c_j)$ in Eq. (\ref{eq:reproj_err}), the geodesic distance $d_{geo}(c_i,c_j)$ can more faithfully measure the dissimilarity between $c_i$ and $c_j$, since the spatial information is taken into account to remove the noises, \ie incorrect correspondences whose homographies happen to be consistent with those of the correct ones. The performance gain of using geodesic distances over reprojection errors will also be evaluated in each of the used dataset in the experiments.


\subsection{Correct Matching Identification by One-class SVM}

At this stage, we apply one-class SVM to distinguishing the correct correspondences from candidate set ${\cal C}$. One-class SVM is one of the state-of-the-art methodologies for unsupervised classification. It separates positive and negative data in an asymmetrical scenario: it assumes that positive data are similar to each other, while negative data are different in their own ways.

In our case, the correct correspondences are usually geometrically and spatially consistent with each other, \ie short geodesic distances among them. On the other hand, the wrong matchings are caused by various factors, so their homographies often irregularly distribute. It results in that the homography of a wrong correspondence tends to be dissimilar to most homographies of all the other correspondences. Thus, our case closely meets the scenario of one-class SVM: the correct and incorrect matchings respectively correspond to positive and negative data. For the set of correspondence candidates, ${\cal C} = \{c_1, ..., c_N\}$, one-class SVM predicts their labels by solving the following constrained optimization problem
\begin{align}
\min_{\bw, \{\epsilon_i\}}\;\;& \frac{1}{2}||\bw||^2 + \frac{1}{Co \cdot N} \sum_{i=1}^N \epsilon_i - \nu   \label{eqn:oneclassSVM} \\%
\text{subject to}\;\;& \bw^{\top} \phi(c_i) \geq \nu - \epsilon_i, \mbox{ for } 1 \leq i \leq N, \nonumber\\%
\;\;& \epsilon_i \geq 0, \mbox{ for } 1 \leq i \leq N, \nonumber
\end{align}
where $Co$ and $\nu$ are the two parameters of one-class SVM. We tuned and fixed them as $Co = 1$ and $\nu = 0.5$ for all the experiments. As a kernel machine, one-class SVM can work on nonlinearly mapped data. The function $\phi$ maps the data, correspondences here, to a {\em Reproduced Kernel Hilbert Space} (RKHS), which is implicitly defined by the adopted kernel. The optimization of one-class SVM can be accomplished by referencing only the inner products of pairs of the mapped data, and the inner product can be efficiently computed via the {\em kernel trick}, \ie $k(c_i,c_j) = \langle \phi(c_i), \phi(c_j) \rangle$.

Note that we don't explicitly define the feature representation of correspondences in ${\cal C}$, but their pair-wise dissimilarity through $d_{geo}$. A kernel function is used to encode the similarity among data. In this work, the kernel matrix $K \in \bbR^{N \times N}$ and the kernel function $k(\cdot,\cdot)$ are defined as follows:
\begin{flalign}
K(i,j) & = k(c_i,c_j)\nonumber\\
& = \exp{(-\frac{d_{geo}^2(c_i,c_j)}{\sigma^2})}, \mbox{ for } 1 \leq i,j \leq N,
\end{flalign}
where $\sigma$ is the hyperparameter. We set $\sigma$ as twice of the average geodesic distance from each correspondence to its nearest neighbor. It follows that each correspondence $c_i$ is predicted via $\mbox{sign}(f(c_i)),$ where score $f(c_i)$ is in form of:
\begin{equation}
 f(c_i) = \bw^{\top} \phi(c_i) - \nu =  \sum_{j=1}^N \alpha_j k(c_i,c_j) - \nu, \label{eqn:prediction}
\end{equation}
and $\{\alpha_j\}_{j=1}^N$ are the optimized coefficients of the support vectors. Note that the results of image matching are often jointly measured by {\em precision} and {\em recall}. For each feature point $u_i^P$ in image $I^P$, we pick its correspondence as the one that has the highest prediction score in ${\cal C}_i$ (cf. Eq. (\ref{eq:C})). All picked correspondences are further sorted according to their prediction scores. Precision-recall analysis can then be carried out with the sorted list and a set of thresholds.

Our approach is featured with its high degree of flexibility in matching images with multiple descriptors. First, any elliptical interest region detectors or their combination can be used to locate feature points. Second, any region descriptors can be applied and fused. No assumption about their feature representations, dimensions, the associated similarity measures has been made. Third, motivated by the observation that the optimal descriptor for matching is image-dependent or even region-dependent, our approach allows each feature point to select its correspondence obtained by any descriptor only if the resulting homography is geometrically and spatially consistent. Besides, our approach is easy to implement. The geodesic distances can be computed via {\em Dijkstra's algorithm} with an approximate version to only update $200$ times on each distance for computational efficiency, and a few packages of one-class SVM are available, such as {\em LibSVM}~\citep{Chang01} which is adopted in this work. The time complexity of our method is between $O(N^2)$ to $O(N^3)$, \ie the complexity of one-class SVM, where $N$ is the number of correspondences. Note that in this work, $N$ grows linearly with respect to the number of descriptors used for fusion.


\section{Experimental Setup\label{sec:imp}}

In this section, we introduce the details of our experimental setting, including the used feature detector, descriptors, baselines, datasets, and evaluation criteria.

\subsection{Feature Detector and Feature Descriptors}

The Hessian affine invariant detector~\citep{Mikolajczyk04} is used in our experiments to detect interest points and their surrounding elliptical support regions. Each detected region is normalized and rotated to the principal orientation. We apply all the feature descriptors to the normalized patches except for geometric blur. We will explain it later. The average number of detected interest points in an image is around $10^3$.

In the experiments, we adopt five feature descriptors, because they capture diverse image characteristics and tend to complement each other. They are listed below in bold and in abbreviation:

\textbf{SIFT:} The {\em SIFT}~\citep{Lowe04} descriptor constructs a $3$D histogram on the gradient map, and stacks the $3$D histogram as a $128$-dimensional vector. This descriptor is known to be rotation and scale invariant.

\textbf{LIOP:} The {\em LIOP}~\citep{Wang11} descriptor models the intensity relationship of a patch, and thus is able to handle monotonic intensity changes. The LIOP features are represented by a $144$-dimensional vector.

\textbf{DAISY:} The {\em DAISY}~\citep{Tola10} descriptor is known for its computational efficiency and invariance to viewpoint changes. It employs spatially varied kernels, and puts high weights on the region near the feature center. The dimension of a DAISY descriptor is $136$.

\textbf{RI:} The {\em Raw intensity} (RI) descriptor is represented by the pixel intensities in a raster scan order. It is the most distinctive descriptor among the five descriptors, but is also the most sensitive to noises and variations. We normalize each support region to $31\times31$ pixels. Thus, a $961$-dimensional RI descriptor is constructed.

\textbf{GB:} The {\em geometric blur} (GB) descriptor~\citep{Berg01} is widely used for shape matching. Like DAISY, it uses spatially varied kernels to deal with the possible deformation. To make the GB descriptor more complementary to others, we enlarge each detected support region by three times, and apply it to the enlarged region to capture shape information with a wider range.

Euclidean distance is used to measure the distance between two regions under each descriptor.

\subsection{Baselines}

For performance comparison, we implemented eight baselines of image feature matching. Four of them are the state-of-the-art feature matching algorithms, including {\em spectral matching} (SM)~\citep{Leordeanu05}, {\em agglomerative correspondence clustering} (ACC)~\citep{Cho09}, {\em Hough voting} (HV)~\citep{Chen13}, and {\em vector field consensus} (VFC)~\citep{Ma14}. The other four are the baselines that fuse multiple descriptors, including {\em concatenation} (CAT), {\em concatenation with Hough voting} (CAT+HV), {\em ranking}, and {\em ratio}. The following describes the eight baselines.

\textbf{SM:} \citet{Leordeanu05} employed a graph structure to realize geometric checking. Their method relaxes the formulation of an integer quadratic programming problem for feature matching, and obtains an approximate solution by solving an eigenvalue problem. In our implementation, the affinity matrix among correspondences is computed by using a modified RBF function in which Euclidean distance is replaced with the reprojection error.

\textbf{ACC:} \citet{Cho09} proposed to iteratively group correspondence clusters based on geometric consistency. The reprojection error is used to measure the geometric consistency, as used in~\citep{Cho09}.

\textbf{HV:} \citet{Chen13} conducted Hough transform for every correspondence and verified its geometric coherence with nearby correspondences in the transformation space. The reprojection error is used in kernel density estimation for geometric verification.

\textbf{VFC:} \citet{Ma14} proposed to identify outliers by modeling inliers with a vector field. We used the implementation provided by~\citet{Ma14} in the experiments.

\textbf{CAT:} This baseline fuses multiple descriptors by concatenating all the feature descriptors. Before concatenation, the standard deviation of pair-wise distances between feature points under each descriptor is firstly computed, and each descriptor is normalized by dividing by the standard deviation. The nearest neighbor search is applied to finding the possible matching candidates.

\textbf{CAT+HV:} This baseline uses the initial candidates constructed by baseline CAT as the input of HV. Compared with CAT, CAT+HV additionally realizes geometric checking by Hough voting.

\textbf{Ranking:} In this baseline, we compute the nearest neighbors of all feature points in image $I^p$ with a specific descriptor, and assign the ranks to these points according to their distances to the nearest neighbors. The procedure is repeated for each descriptor. For each feature point in $I^p$, we determine its correspondence based on the descriptor that has the highest rank at this point.

\textbf{Ratio:} For each point in $I^p$, we find its first two nearest neighbors with a specific descriptor, and compute the distance ratio between the first two neighbors. The smaller the ratio is, the more confident the descriptor is at this point. The procedure is repeatedly done for each descriptor. For this feature point, we determine its correspondence by using the descriptor with the smallest distance ratio.

It is worth mentioning that baselines CAT and Ranking consider all the descriptors are equally important, so their feature vectors or ranks can be directly concatenated or compared. However, the performances of these descriptors in fact vary from image to image. The performance of baseline Ratio degrades when there is high variation among the distance distributions of all the descriptors. The reason is that the distance ratios cannot be compared across different descriptors in that situation.

The first four baselines take one descriptor into account at a time, while the other four baselines and our approach consider all the adopted descriptors jointly. For fair comparison, all the baselines and our approach use the same detected interest regions, descriptors, and evaluation criteria in the experiments.

\subsection{Datasets}

The performance of our approach is evaluated on four benchmarks of image matching, including Co-recognition ({\snu} for short) dataset~\citep{Cho08}, {\Rrwm} dataset~\citep{Cho10}, Symfeat ({\sym} for short) dataset~\citep{Hauagge12} and {\vgg} dataset~\citep{Mikolajczyk05b}. To get preeminent performance on all of them is very challenging, because the four datasets contain a variety of variations, such as diverse kinds of deformation, various types of scenes, and different numbers of common objects. They jointly serve as a good test bed for performance evaluation.


\textbf{{\snu} dataset:} It consists of six image pairs. There are multiple common objects in every pair. Large changes in viewpoints, rotations and scales combined with noises coming from the clutter backgrounds and occlusions make matching quite difficult on this dataset.

\textbf{{\Rrwm} dataset:} It gathers $30$ pairs of images. Each pair contains different object instances of the same category. The performance of a descriptor fluctuates a lot between different image pairs of this dataset due to the diversity of intra-class variations.

\textbf{{\sym} dataset:} It is composed of $46$ image pairs of architectural scenes. The dramatic variations of this dataset range from lighting conditions (day/night), ages (old/nowadays scene) to rendering styles (photograph/drawing).

\textbf{{\vgg} dataset:} It contains eight image sets, each of which contains images with five different degrees of a specific type of variation. Total five types of variation are included in the dataset, namely viewpoint changes, scale changes, image blur, JPEG compression, and illumination changes. For each image set, we carry out image matching with two different degrees of variation in the experiment, \ie the first image vs. the second one and the first one vs. the fourth one, which represent the slight and drastic variations in matching, respectively.

\subsection{Evaluation criteria}

For performance measure, the evaluation metrics used in the experiments are introduced. For datasets {\vgg} and {\sym}, we follow~\citet{Mikolajczyk05b}, and consider that a correspondence is correct if the overlap error is less than $50\%$. For datasets {\Rrwm} and {\snu} where manually annotated ground truth is available, a correspondence is correct if the distance between the matched feature point and the ground truth is within eight pixels.

After determining the correctness of correspondences, the performance of a matching algorithm can be measured by jointly taking {\em precision} and {\em recall} into account. While precision is the fraction of detected correspondences that are correct, recall is the fraction of correct correspondences that are detected. Specifically, the two terms are respectively defined as
\begin{equation}
\preci = \frac{\TP}{\TP+\FP}, \label{eqn:precision}
\end{equation}
and
\begin{equation}
\reca = \frac{\TP}{\TP+\FN}, \label{eqn:recall}
\end{equation}
where $\TP$ and $\FP$ are the numbers of correctly and wrongly detected correspondences by a matching method, respectively. $\FN$ is the number of correct correspondences that are not detected.

For each matching approach, including ours and the eight adopted baselines, all the detected correspondences are sorted by its own criterion, such as the element values of the eigenvector in baseline SM, the ratio values in baseline Ratio, and the prediction score, Eq.~(\ref{eqn:prediction}), in our approach. By sampling on the sorted lists, the performance of each approach can then be represented by a {\em precision-recall curve} or {\em mean average precision} (mAP).

\comment{
\begin{table*}[tH] 
	\normalsize
	\begin{center}
		\caption{The characteristics of the four datasets} 	
		\label{table:dataset}
		\tabcolsep=3pt 	\renewcommand{\arraystretch}{1.2}
		\begin{centering} 	 \begin{tabular}{c|cccc} 	\hline \hline
				  & \snu  & \Rrwm  & \sym  & \vgg \\
				\hline
				Number of common objects & Multiple object & Single object &  Single(Scene) & Single(Scene) \\
				
				 & 78.25 & 33.65 & 37.31 & 88.39 \\
				\hline
				
				Performance Gain & 5.02 & 3.60 & 5.26 & 0.14 \\
				\hline
			\end{tabular} 	
		\end{centering}
	\end{center}
\end{table*}
}


\section{Experimental Results\label{sec:exp}}

The performance of the proposed approach is evaluated and analyzed in this section. Totally, three sets of experiments are conducted. First, the transformation space is visualized to verify that correct correspondences established by all the descriptors gather together in that space, while incorrect ones distribute irregularly. We also visualize the homography spaces when the reporjection error and the proposed geodesic distance serve as the distance functions, respectively. Second, our approach is compared with the eight baselines on the four benchmarks of feature matching. The obtained quantitative results of all methods are presented in the form of mAPs and precision-recall curves. Third, we show the matching results to demonstrate that our approach can leverage multiple, complementary descriptors to achieve remarkable performance improvement in feature matching.

\subsection{Homography Space Visualization}

To check whether the homography space is qualified to serve as a uniform domain for descriptor fusion, we visualize it by {\em multi-dimensional scaling} (MDS)~\citep{Cox94}, which can approximate the pair-wise distances between data in a $2$D space.

\begin{figure}[tH]
\begin{center}
\hspace*{-0.22 cm}\begin{tabular}{cc}
        \includegraphics[width = 1.6 in]{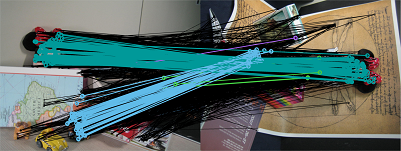} & \hspace{-0.4 cm} \includegraphics[width = 1.6 in]{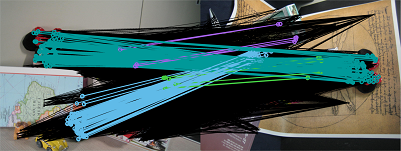}\\
         (a) & \hspace{-0.4cm}(b) \\[5pt]
       \fbox{\includegraphics[height = 1.4 in, width = 1.5 in]{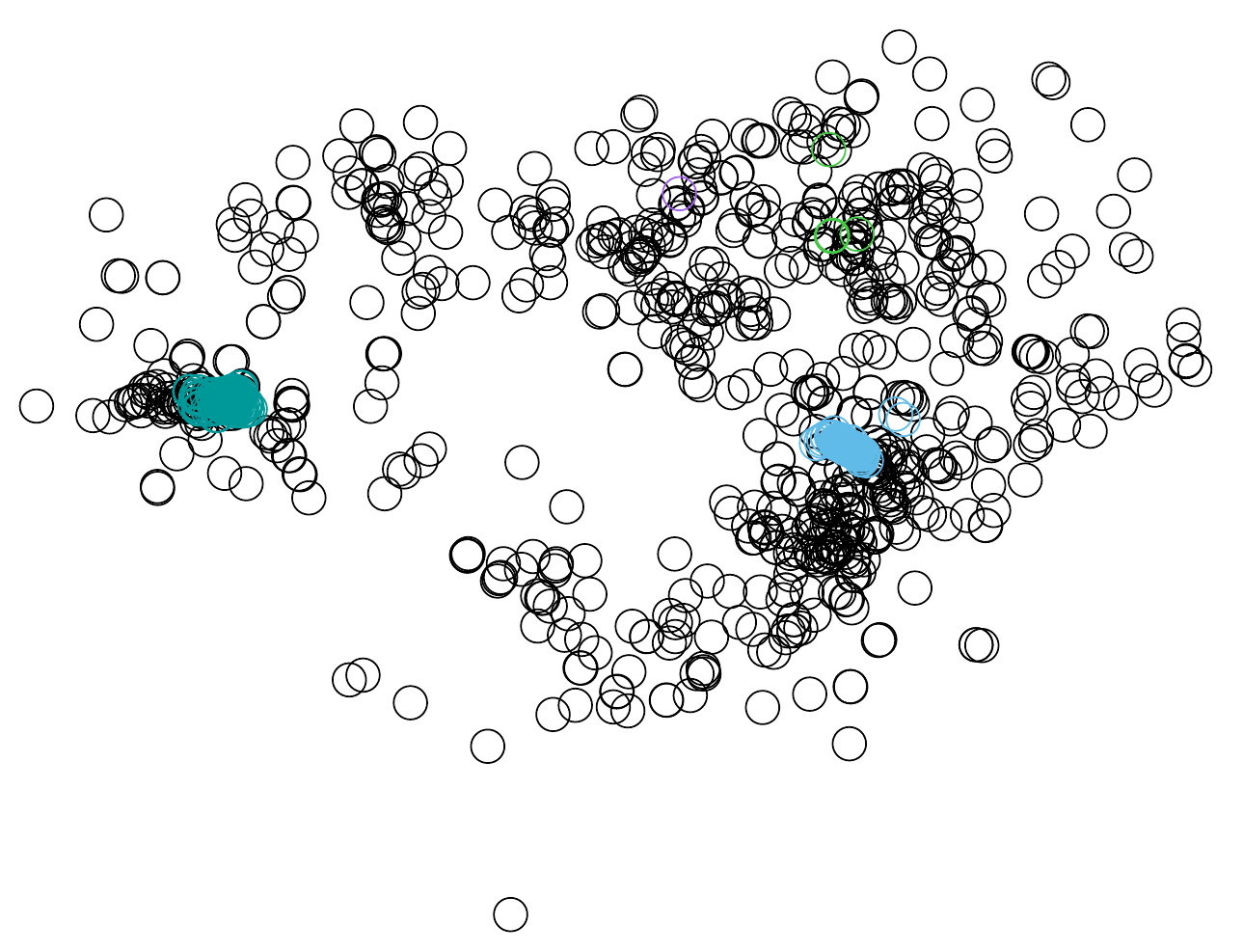}} &\hspace{-0.4cm}
       \fbox{\includegraphics[height = 1.4 in, width = 1.5 in]{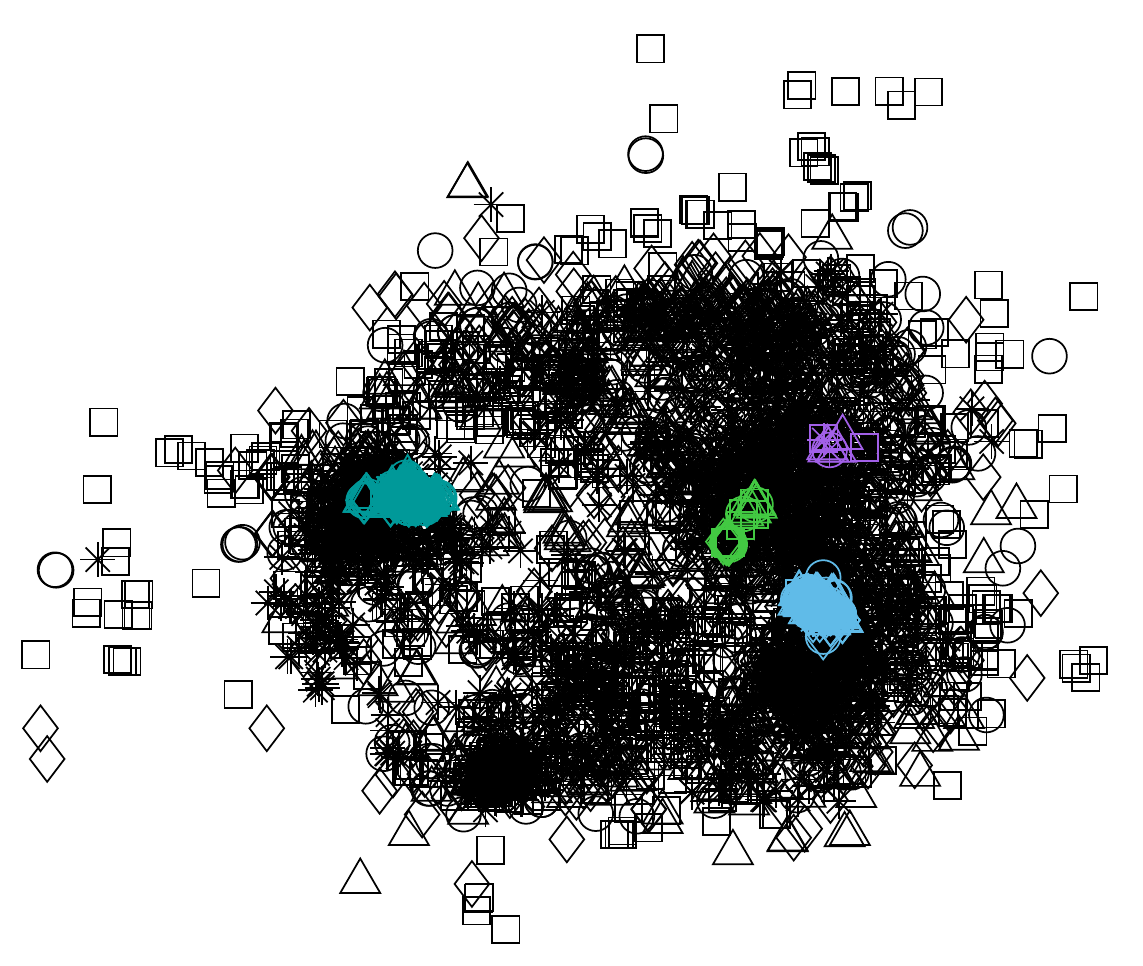}}\\[2pt]
        (c) & \hspace{-0.4cm}(d)
\end{tabular}
\end{center}
 \vspace{-0.1 in}
\caption{(a) The matching candidates given by SIFT on image pair {\tt Minnies} of the {\snu} dataset. (b) The matching candidates given by all the five descriptors on the same image pair. (c) $2$D visualization of the homograph space of the matchings in (a). (d) $2$D visualization of the homograph space of the matchings in (b).}
\label{fig:MDS_SIFT_Fusion}
\end{figure}

\figname~\ref{fig:MDS_SIFT_Fusion}(a) and \figname~\ref{fig:MDS_SIFT_Fusion}(b) show the sets of the initial matching candidates, via Eq.~(\ref{eq:C}), by using SIFT and by using all the five descriptors, respectively. There are four common objects in this pair of images. The correct matchings are colored according to the objects that they lie on. By using the geodesic distance presented in {\secname}\ref{sec:our-geo}, the two sets of matchings are visualized in \figname~\ref{fig:MDS_SIFT_Fusion}(c) and \figname~\ref{fig:MDS_SIFT_Fusion}(d), respectively. It turns out that no matter how many descriptors are used, the correct (colored) correspondences gather together, while wrong matchings distribute irregularly. This example also points out that using multiple descriptors helps to find out the correct correspondences that are sparsely detected with a single descriptor and may be neglected, such as the purple correspondences in \figname~\ref{fig:MDS_SIFT_Fusion}(c). In \figname~\ref{fig:MDS_SIFT_Fusion}(d), the purple correspondences given by diverse descriptors can mutually support in both geometric and spatial coherence estimation. It implies that they are more probably predicted as correct correspondences by one-class SVM.

\begin{figure}[tH]
\begin{center}
\hspace*{-0.22 cm}
\begin{tabular}{cc}

\includegraphics[width = 1.6 in]{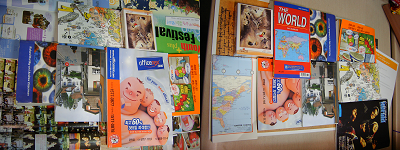} & \hspace{-0.3 cm}\includegraphics[width = 1.6 in]{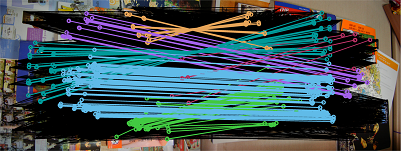}\\
         (a) & \hspace{-0.4 cm}(b) \\[5pt]
       \fbox{\includegraphics[height = 1.4 in, width = 1.5 in]{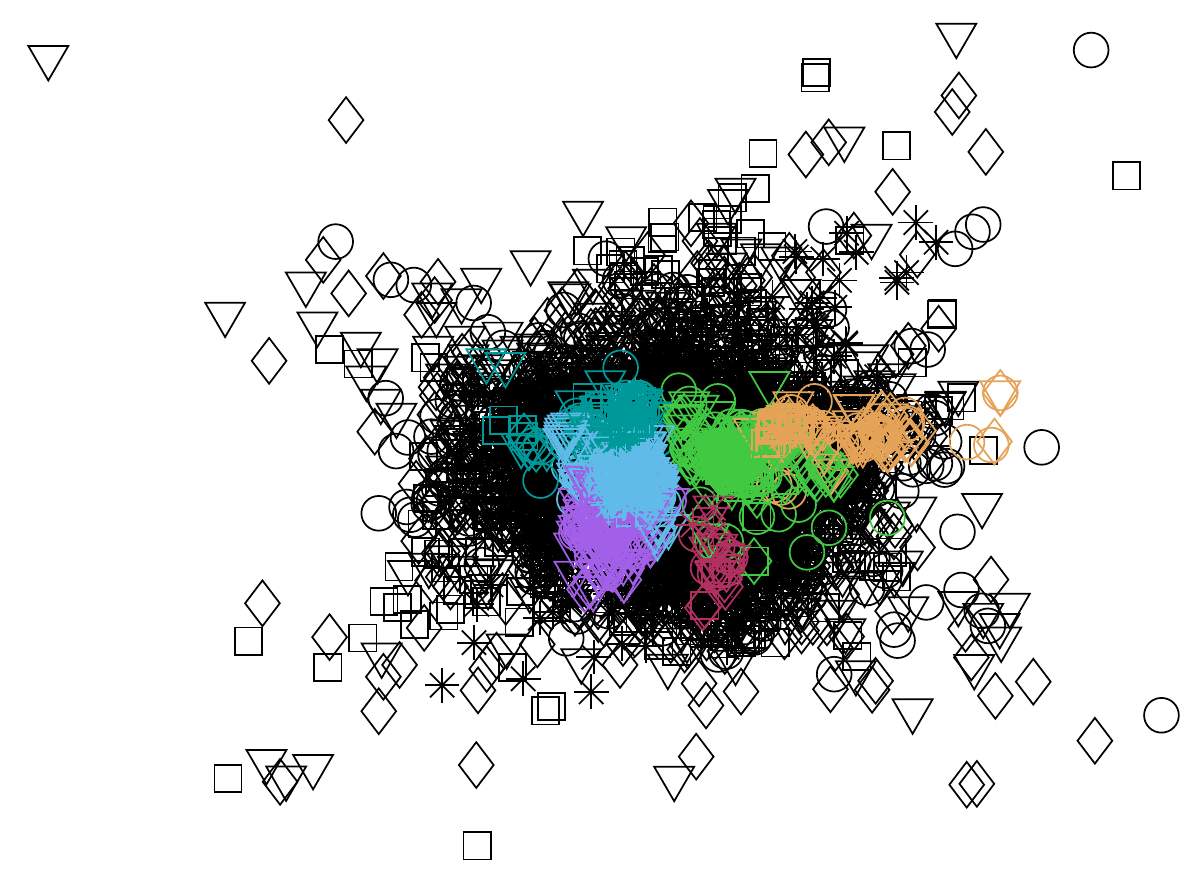}} &\hspace{-0.4 cm}
       \fbox{\includegraphics[height = 1.4 in, width = 1.5 in]{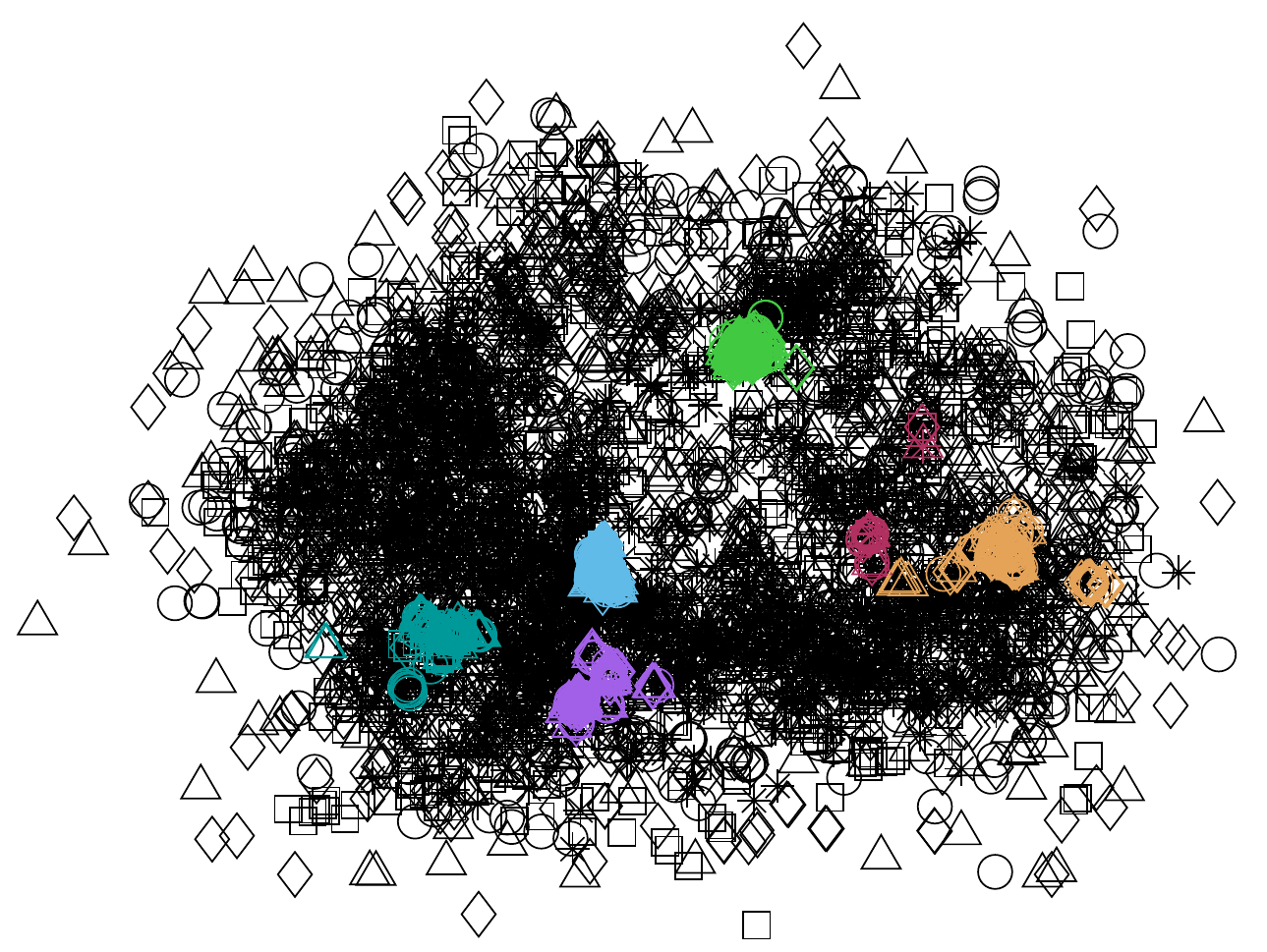}}\\[2pt]
        (c) & \hspace{-0.4 cm}(d)
\end{tabular}
\end{center}
 \vspace{-0.1 in}
\caption{(a) An image pair, {\tt Books} of the {\snu} dataset. (b) The initial correspondences. (c) $2$D visualization of these correspondences in the homography space when the reprojection error is used. (d) $2$D visualization of these correspondences in the homography space when the developed geodesic distance is used.}
\label{fig:MDS_Rep_Geo}
\end{figure}

As described in {\secname}\ref{sec:our-geo}, the developed geodesic distance computed over the designed graph takes both geometric consistency and spatial continuity into account. We compare the geodesic distance with the reprojection error by visualizing the homography spaces that they induce. In~\figname~\ref{fig:MDS_Rep_Geo}(a), two images to be matched are shown. The initial correspondence candidates are given in~\figname~\ref{fig:MDS_Rep_Geo}(b). We calculate the dissimilarity between these correspondences by using the reprojection error and the geodesic distance, and show their distributions in the homography space in~\figname~\ref{fig:MDS_Rep_Geo}(c) and \ref{fig:MDS_Rep_Geo}(d), respectively. It can be observed in~\figname~\ref{fig:MDS_Rep_Geo}(c) that the correct correspondences on an object may mix with the correct correspondences on the other objects. On the other hand, the correct correspondences on an object tightly assemble in~\figname~\ref{fig:MDS_Rep_Geo}(d), and are well separated from the rest. It implies that the developed geodesic distance can effectively consider both geometric and spatial consistency to better discover object-aware homographies. This property facilitates correct correspondence identification by one-class SVM.

\subsection{Quantitative Results}

\begin{table}[tH] 
	\normalsize
	\begin{center}
		\caption{The accuracy in mAP of using the reprojection error and the geodesic distance in our approach} 	
		\label{table:reprojection_geodesic}
		\tabcolsep=3pt 	\renewcommand{\arraystretch}{1.2}
\small{
		\begin{centering} 	 \begin{tabular}{c|cccc} 	 \hline \hline
				mAP ($ \% $) & \snu  & \Rrwm  & \sym  & \vgg \\
				\hline
				Reprojection Error & 73.13 & 35.97 & 32.05 & 88.25 \\
				
				Geodesic Distance & 78.46 & 40.88 & 37.31 & 88.39 \\
				\hline\hline
				
			\comment{	Performance Gain & 5.33 & 4.91 & 5.26 & 0.14 \\
				\hline}
			\end{tabular} 	
		\end{centering}}
	\end{center}
\end{table}

Following the previous discussion, we evaluate the performance of using the proposed geodesic distance and using the reprojection error in our approach, and report the accuracies in mAP in~\tabname~\ref{table:reprojection_geodesic}. The mAPs on the four datasets are improved when the geodesic distance is used. It indicates that the geodesic distance more faithfully grasps the intrinsic relationships between correspondences by exploring the graph which encodes both the spatial and geometric consistency. The performance gains, about $5\%$, on the first three datasets, \ie \snu, \Rrwm, and \sym, are remarkable. The main reason is that the multiple common objects in~\snu~and the foregrounds and backgrounds in~\Rrwm~and \sym~typically have diverse transformations in matching, but each of these transformations tends to vary smoothly in the spatial domain. Hence, modeling spatial coherence is helpful. On the other hand, all interest points in each image of dataset \vgg~almost undergo the same transformation in matching. Giving additional spatial information does not help much in the cases.

\begin{table}[tH] 
	\normalsize
	\begin{center}
		\caption{The performances in mAP of the eight baselines and our approach on the four datasets} 	
		\label{table:map}
		\tabcolsep=3.5pt 	\renewcommand{\arraystretch}{1.3}
        \small{
		\begin{centering} 	 \begin{tabular}{llllll} 	\hline \hline
				\raisebox{-2ex}[0cm][0cm]{method} & \raisebox{-2ex}[0cm][0cm]{descriptor} & \multicolumn{4}{c}{dataset}\\
				\cline{3-6}
				& & \snu  & \Rrwm  & \sym  & \vgg \\
				\hline\hline
				& {SIFT} & 47.59 & 21.90 &  16.31 & 60.93 \\
				& {LIOP} & 21.16 & 14.61 &  15.96 & 70.06 \\
				SM & DAISY & 35.76 & 20.91 &  17.85 & 71.43 \\
				& {RI} & 14.61 & 10.37 &  15.35 & 67.60 \\
				& {GB} & 9.75 & 14.18 &  16.91 & 65.57 \\
				\cline{2-6}
				& Average & 25.77 & 16.40 &  16.48 & 67.12 \\
				\hline
				& {SIFT} & \textit{60.28}* & 21.81 &  \textit{21.73}* & 77.96 \\
				& {LIOP} & 29.83 & 7.41 &  19.27 & 81.92 \\
				ACC & DAISY & 36.49 & 17.88 &  20.85 & 81.63 \\
				& RI & 15.10 & 7.51 &  18.21 & 77.29 \\
				& GB & 8.88 & 16.16 &  19.53 & 74.02 \\
				\cline{2-6}
				& Average & 30.12 & 14.15 &  19.92 & 78.56 \\
				\hline
				& SIFT & 60.12 & 23.73 &  18.73 & 77.95 \\
				& LIOP & 43.97 & 15.44 &  17.28 & 82.41 \\
				HV & DAISY & 50.06 & 22.32 &  18.84 & 82.74 \\
				& RI & 37.14 & 14.55 &  16.78 & 78.69 \\
				& GB & 12.08 & \textit{30.24}* &  19.18 & 76.40 \\
				\cline{2-6}
				& Average & 40.67 & 21.25 &  18.16 & 79.64 \\
				\hline
				& SIFT & 31.11 & 21.44 &  18.77 & 79.67 \\
				& LIOP & 11.79 & 9.26 &  15.25 & \textit{84.24}* \\
				VFC & DAISY & 16.29 & 18.74 &  16.37 & 83.40 \\
				& RI & 4.51 & 4.46 &  13.40 & 74.54 \\
				& GB & 1.77 & 21.60 &  15.42 & 72.19 \\
				\cline{2-6}				
				& Average & 13.09 & 15.10 &  15.84 & 78.81 \\
				\hline
				CAT & All & 19.62 & 6.20 &  7.22 & 58.76 \\
				\hline
				CAT+HV & All & 39.70 & 15.75 &  11.90 & 70.62 \\
				\hline				
				Ranking & All & 48.48 & 19.38 &  23.00 & 82.05 \\
				\hline				
				Ratio & All & 53.61 & 18.62 &  23.62 & 85.07 \\
				\hline				
				DE (Ours) & All & \textbf{78.46} & \textbf{40.88} &  \textbf{37.31} & \textbf{88.39} \\
				\hline\hline
			\end{tabular} 	
		\end{centering}}
	\end{center}
\end{table}

We evaluate and compare our approach, {\em Descriptor Ensemble} or DE for short, with the eight baselines. Five different descriptors, including SIFT, LIOP, DAISY, RI, and GB, are considered. \tabname~\ref{table:map} summarizes the performances in mAP of all the approaches on the four benchmarks. The first four approaches to image matching, \ie SM~\citep{Leordeanu05}, ACC~\citep{Cho09}, HV~\citep{Chen13}, VFC~\citep{Ma14}, consider a single descriptor at a time. Their performances with each of the five descriptors as well as the average performances are reported. The other four baselines and our approach can jointly take multiple descriptors into account, so only the performances of descriptor fusion are reported. In \tabname~\ref{table:map}, the best performance on each benchmark is given in bold, while the best performance by using a single descriptor is given in italic and comes with a star sign.

\begin{figure*}[tH]
\begin{center}
\hspace*{-0.1 cm}\begin{tabular}{cccc}
\includegraphics[width = 1.65 in]{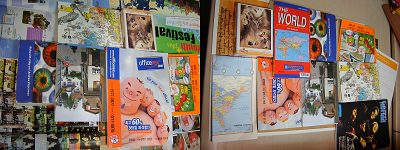} & \hspace{-0.4 cm}
\includegraphics[width = 1.65 in]{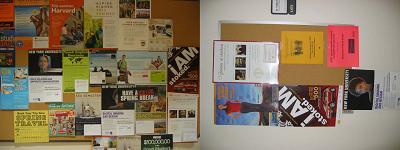} & \hspace{- 0.4 cm}
\includegraphics[width = 1.65 in]{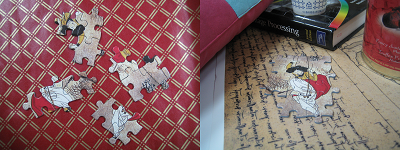} & \hspace{-0.4 cm}
\includegraphics[width = 1.65 in]{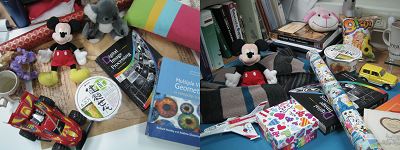} \\
\includegraphics[width = 1.6 in]{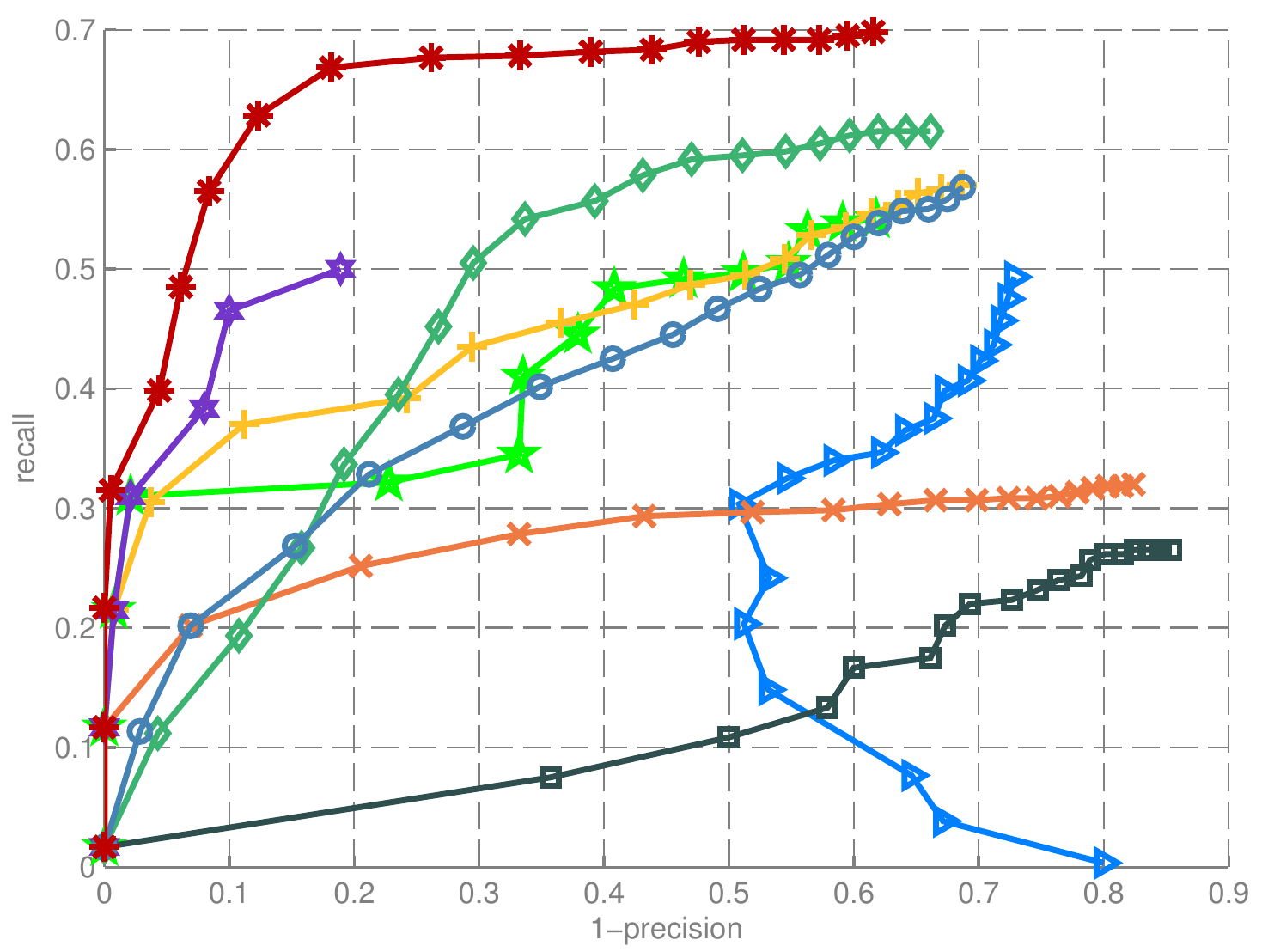} &\hspace{-0.4 cm}
\includegraphics[width = 1.6 in]{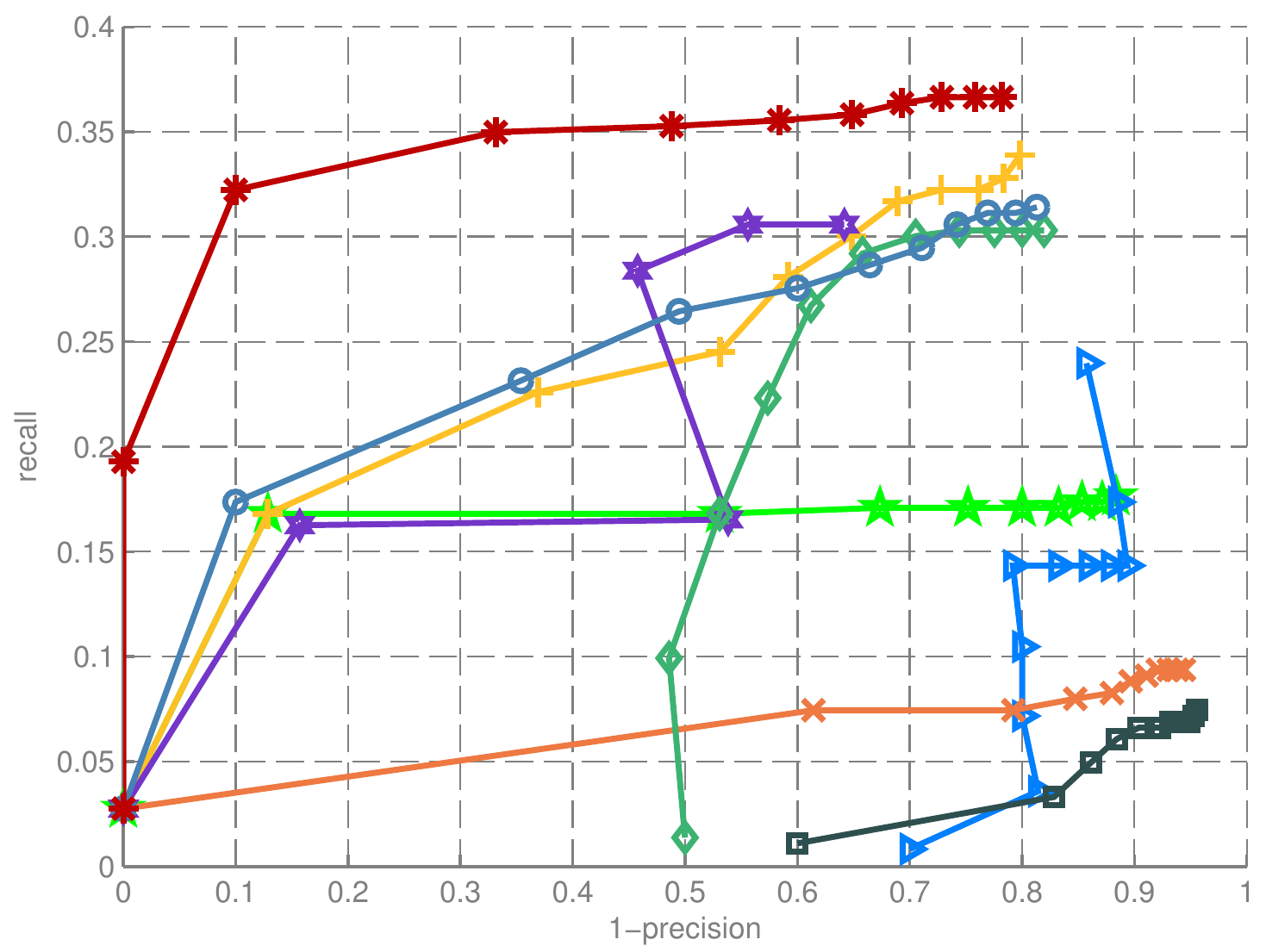} &\hspace{- 0.4 cm}
\includegraphics[width = 1.6 in]{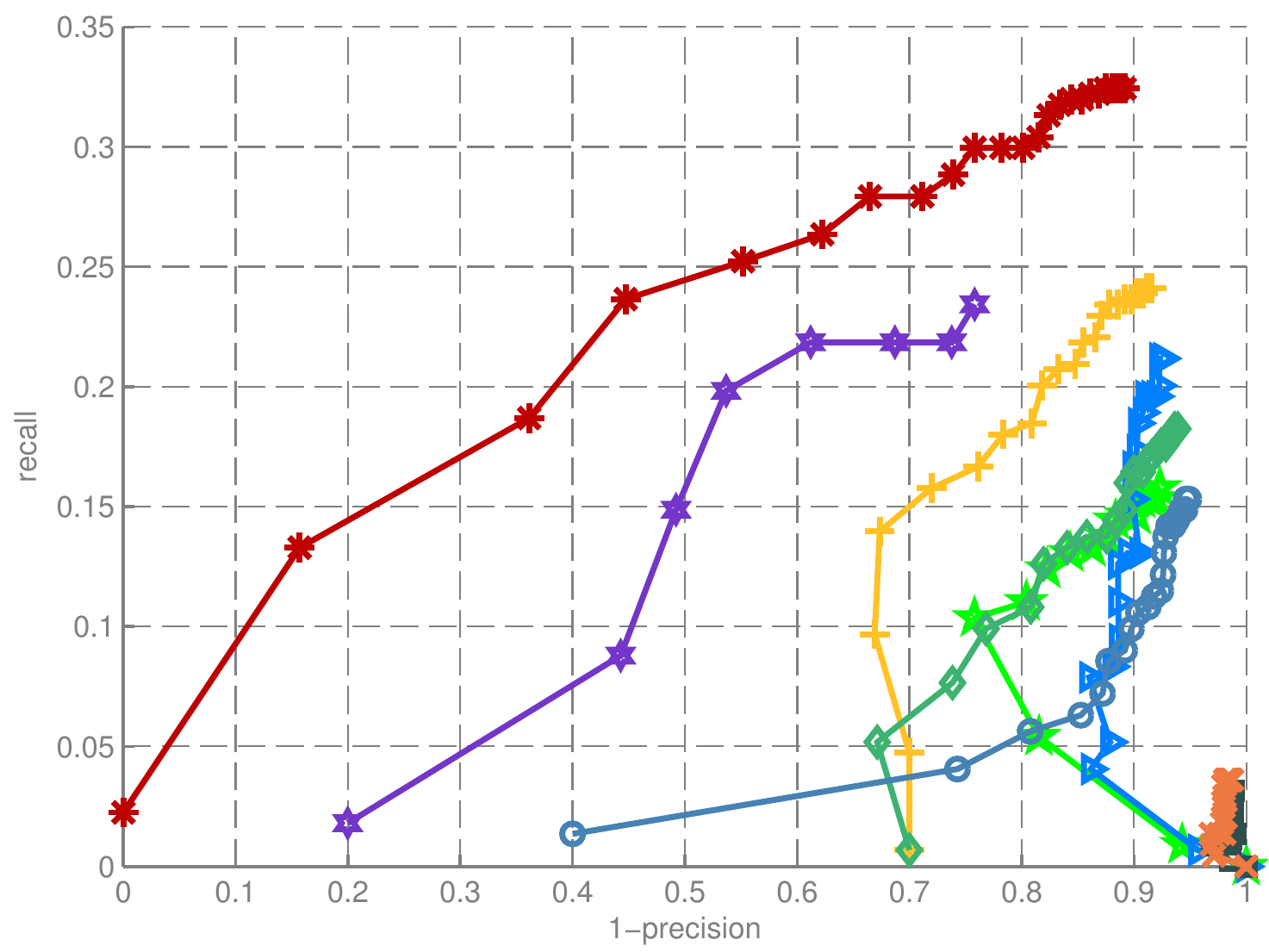} &\hspace{-0.4 cm}
\includegraphics[width = 1.6 in]{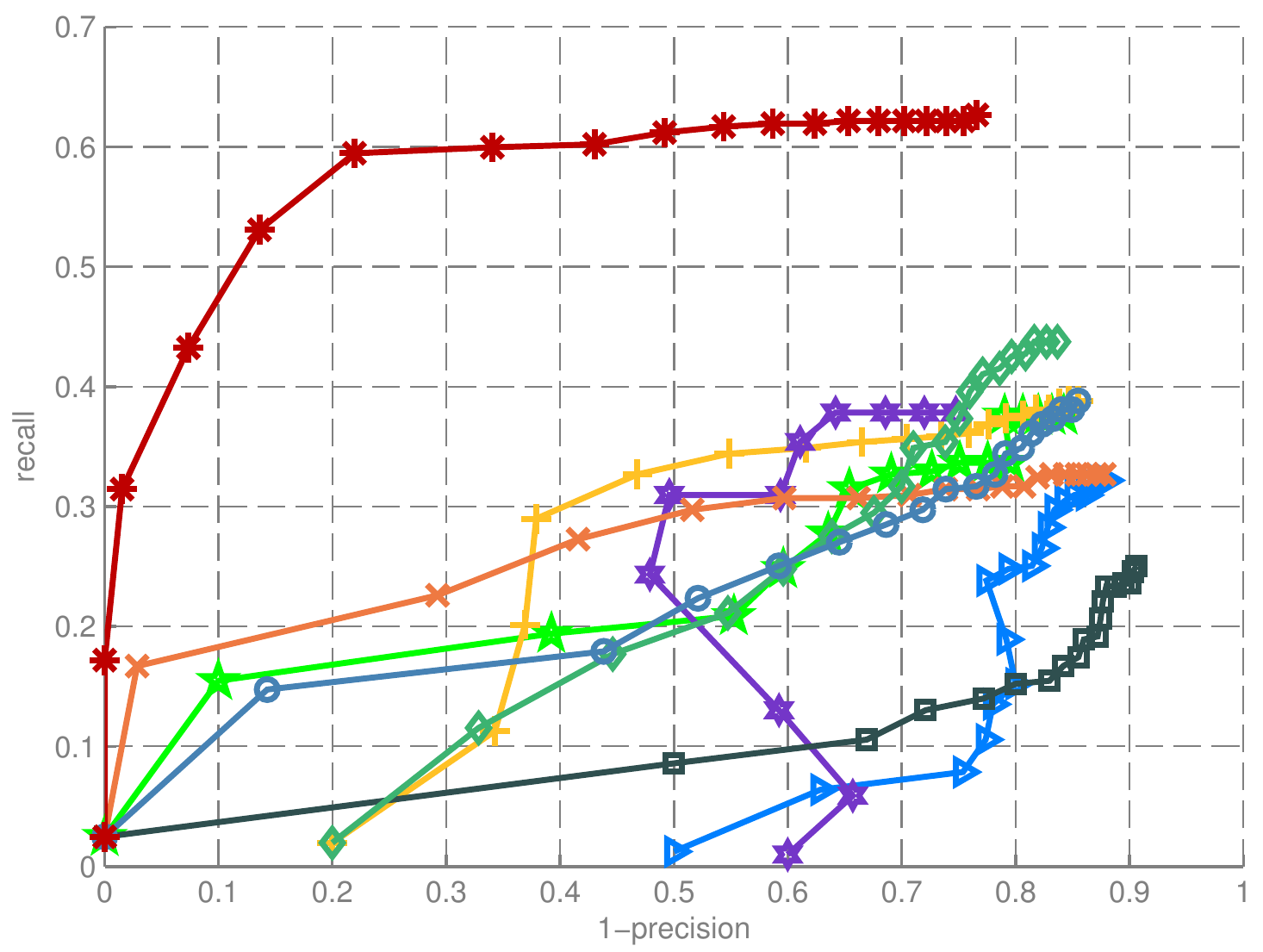} \vspace{-0.05in}\\%
(a) \tt{ Books} & \hspace{-0.4 cm}(b) \tt{ Bulletins} & \hspace{-0.4 cm}(c) \tt{ Jigsaws} & \hspace{-0.4 cm}(d) \tt{ Mickeys} \vspace{0.05in}\\%
\includegraphics[width = 1.6 in]{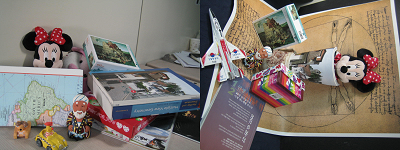} &\hspace{-0.4 cm}
\includegraphics[width = 1.6 in]{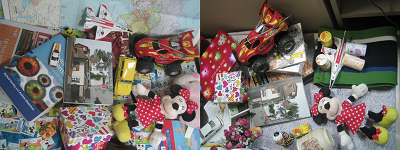} &\hspace{-0.4 cm}
\includegraphics[width = 1.6 in]{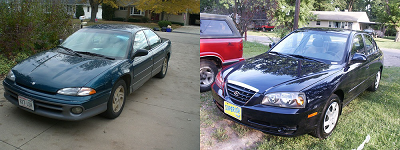} &\hspace{-0.4 cm}
\includegraphics[width = 1.6 in]{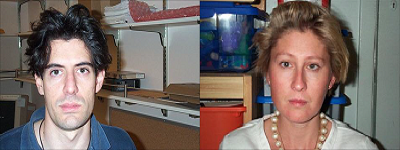} \\
\includegraphics[width = 1.6 in]{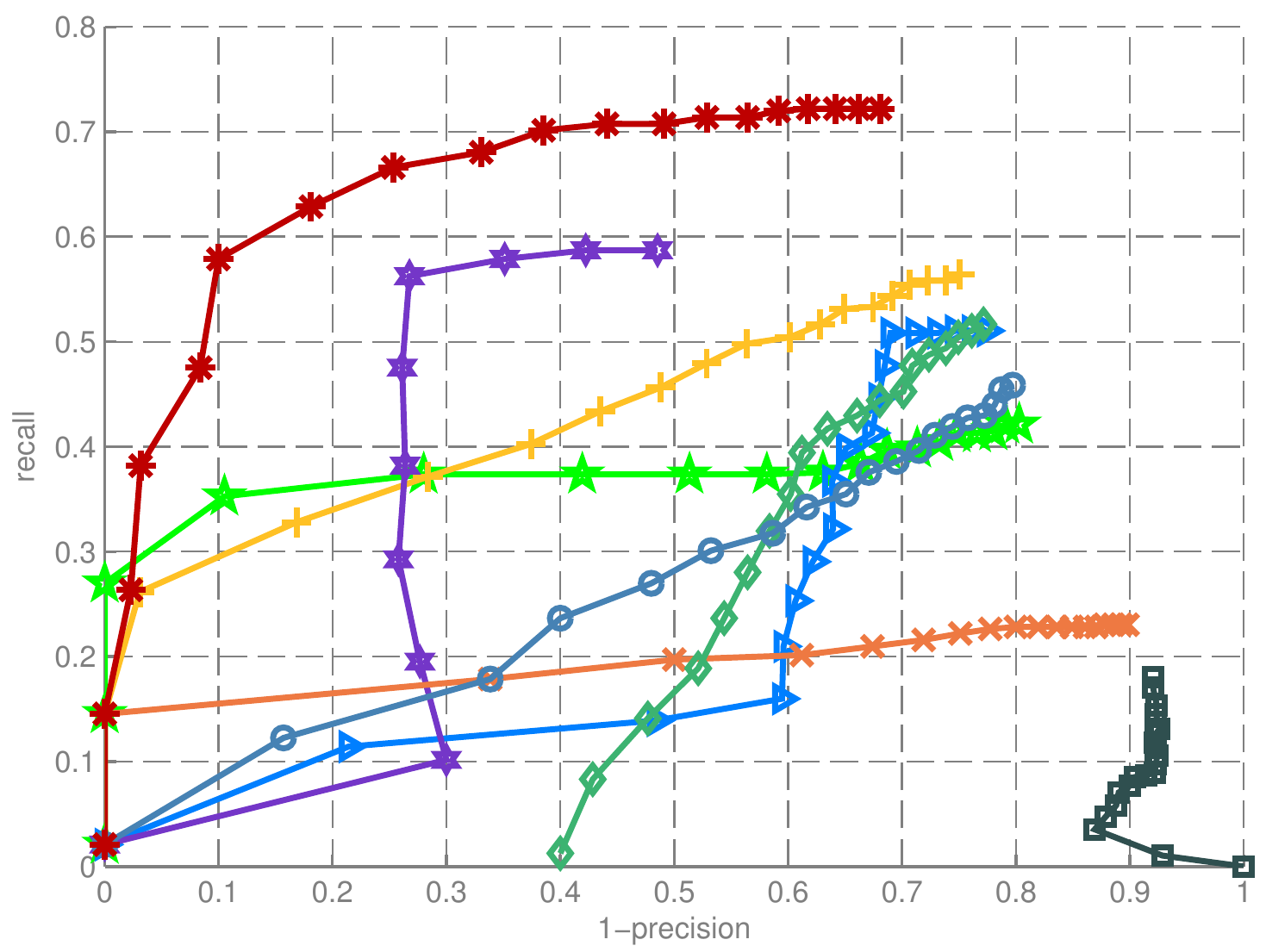} &\hspace{-0.4 cm}
\includegraphics[width = 1.6 in]{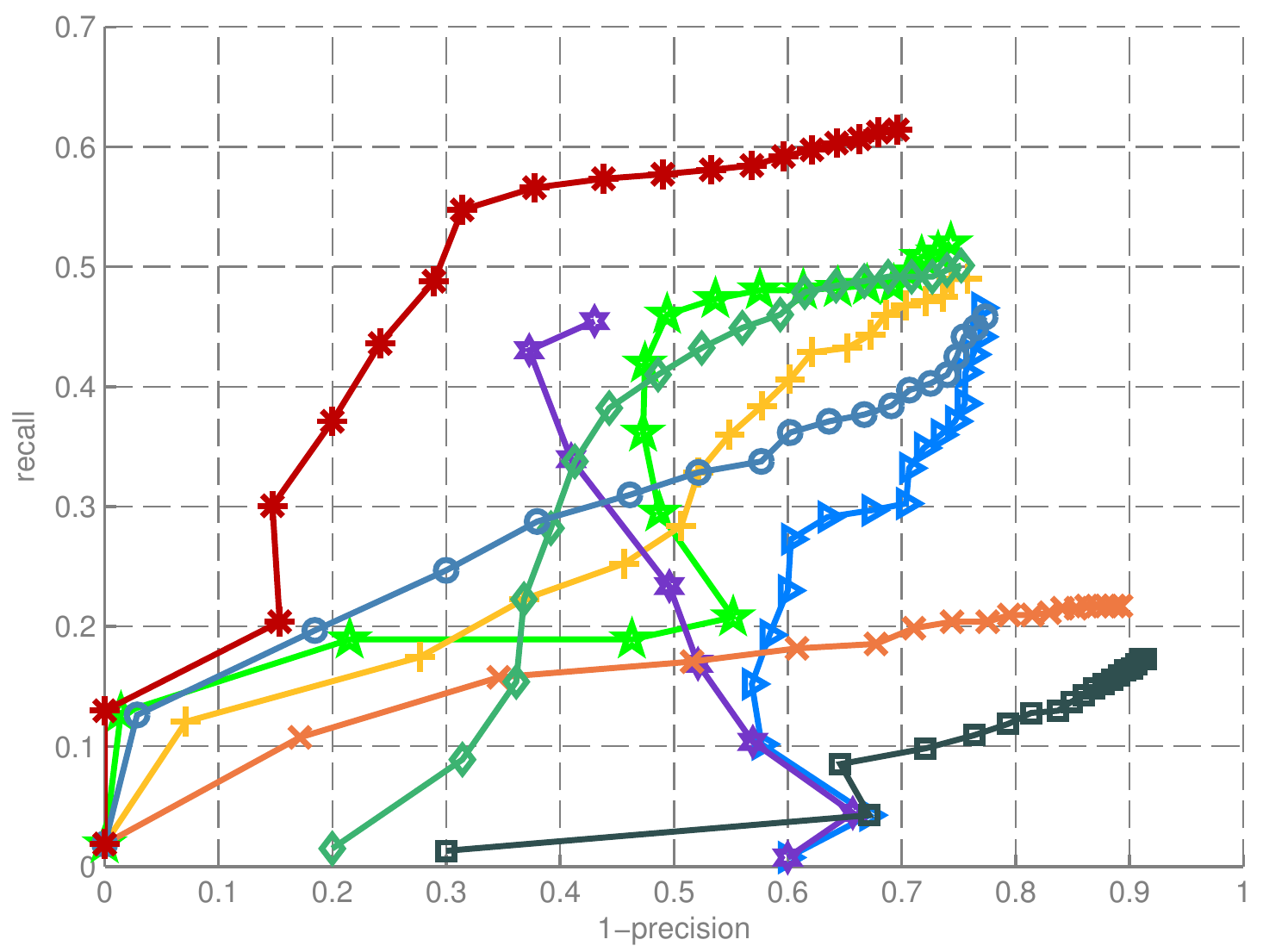} &\hspace{-0.4 cm}
\includegraphics[width = 1.6 in]{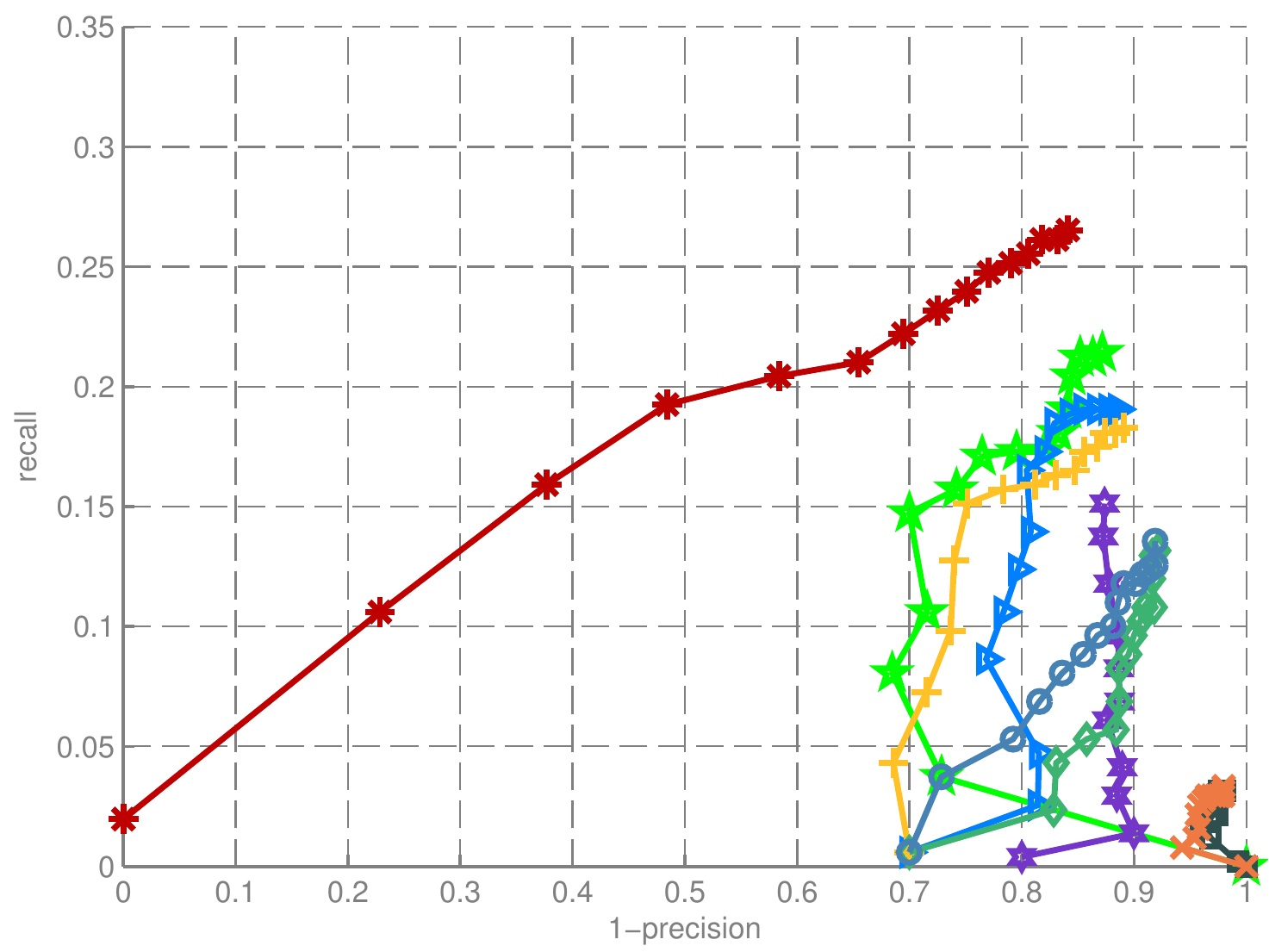} &\hspace{-0.4 cm}
\includegraphics[width = 1.6 in]{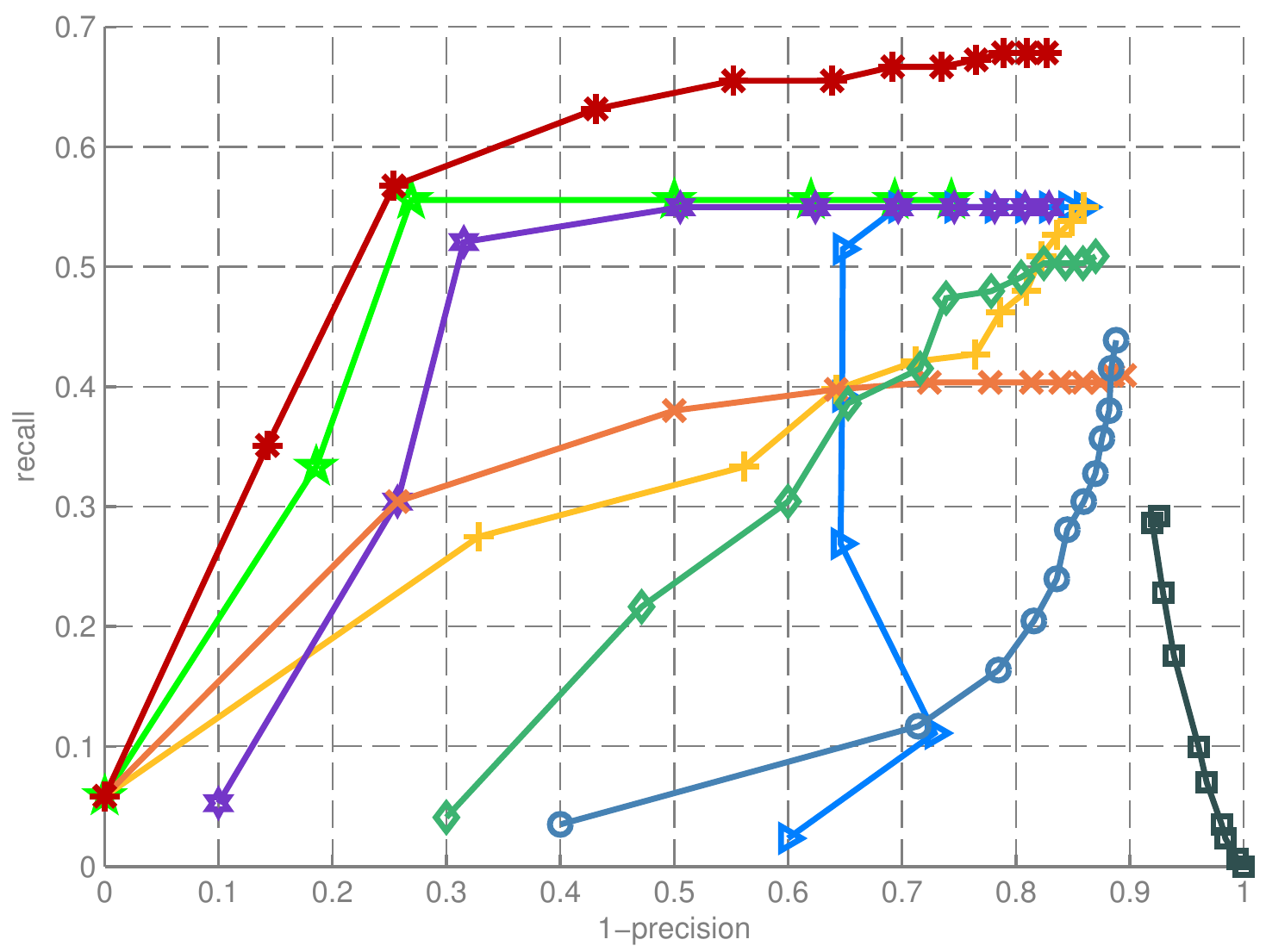} \vspace{-0.05in}\\%
(e) \tt{ Minnies} & \hspace{-0.4 cm}(f) \tt{ Toys} & \hspace{-0.4 cm}(g) \tt{ cars} & \hspace{-0.4 cm}(h) \tt{ face} \vspace{0.05in}\\%
\includegraphics[width = 1.6 in]{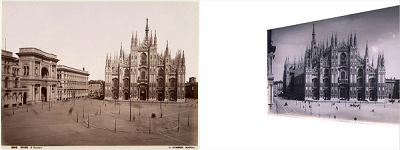} &\hspace{-0.4 cm}
\includegraphics[width = 1.6 in]{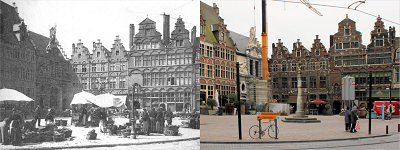} &\hspace{-0.4 cm}
\includegraphics[width = 1.6 in]{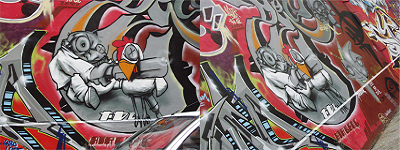} &\hspace{-0.4 cm}
\includegraphics[width = 1.6 in]{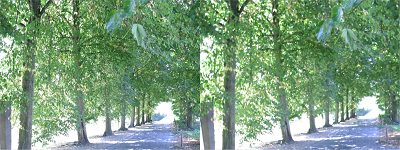} \\
\includegraphics[width = 1.6 in]{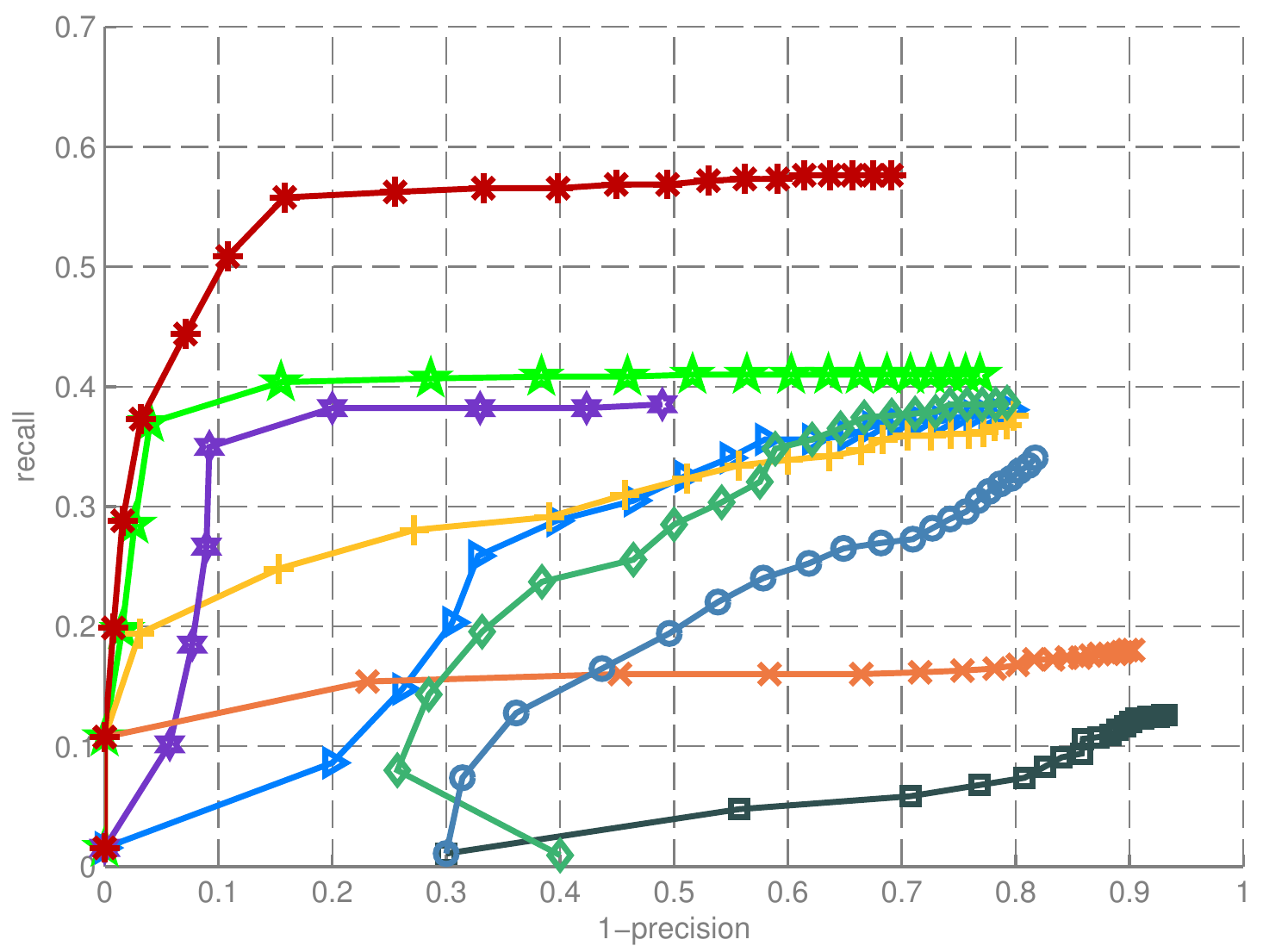} &\hspace{-0.4 cm}
\includegraphics[width = 1.6 in]{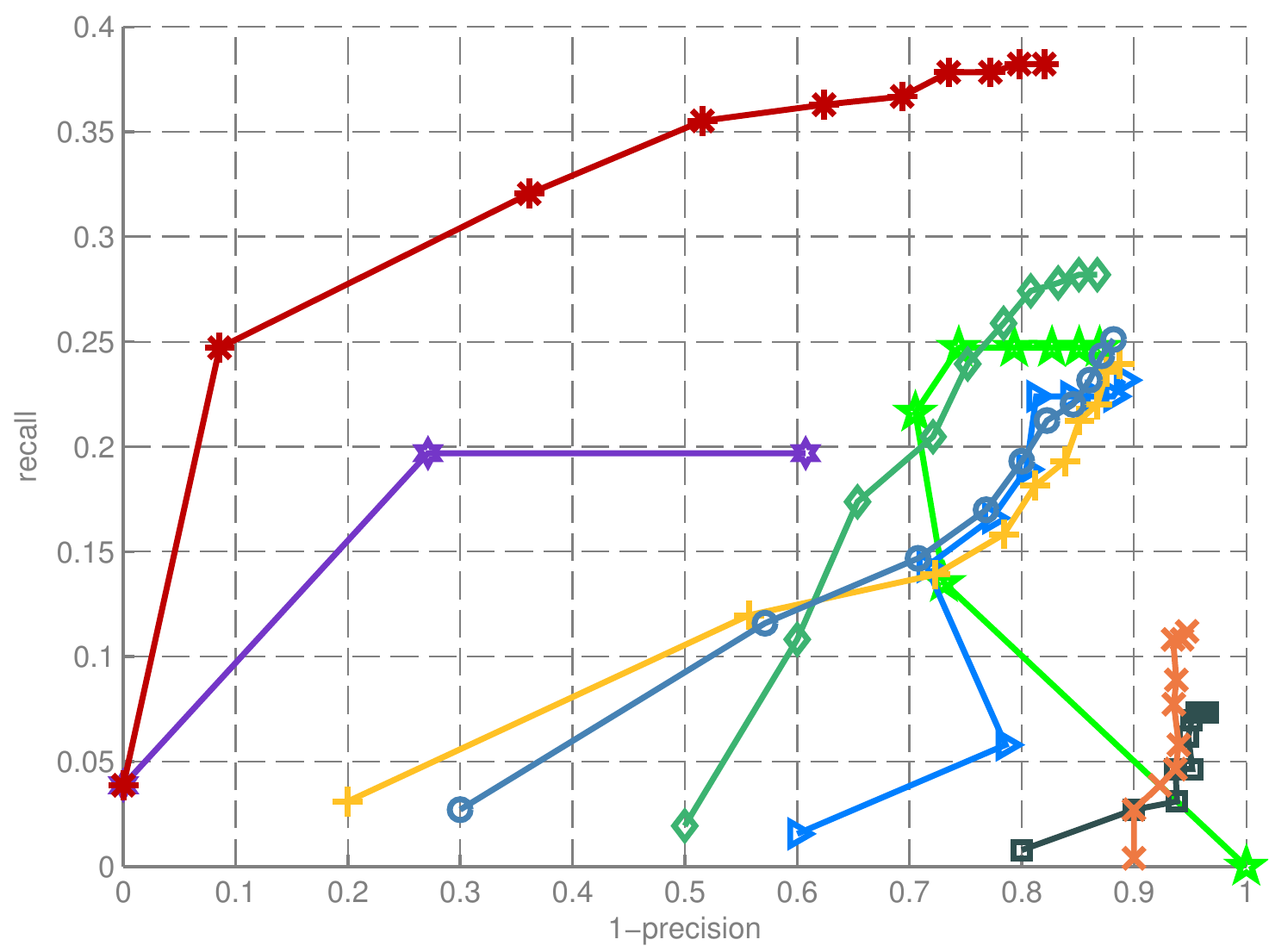} &\hspace{-0.4 cm}
\includegraphics[width = 1.6 in]{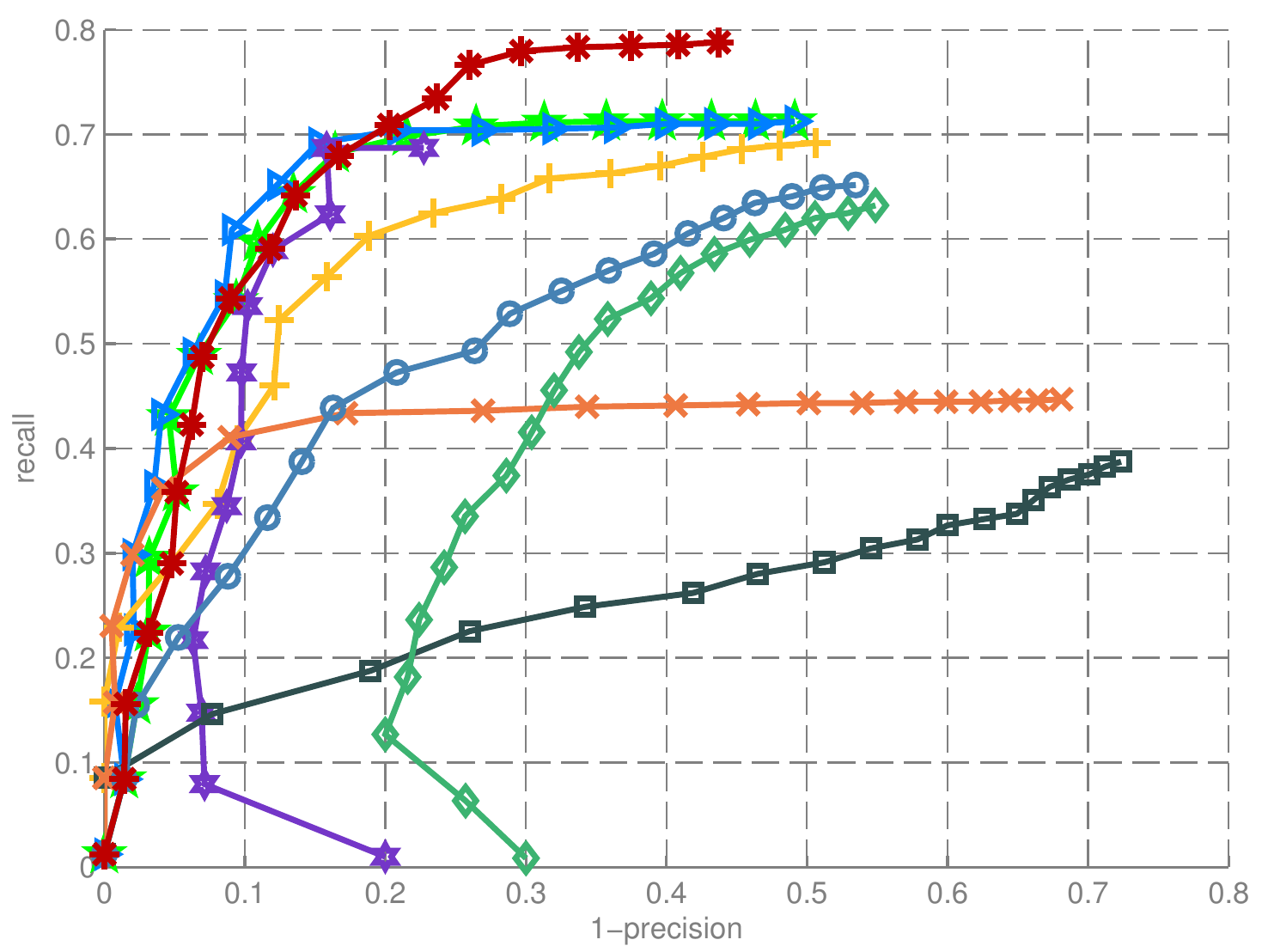} &\hspace{-0.4 cm}
\includegraphics[width = 1.6 in]{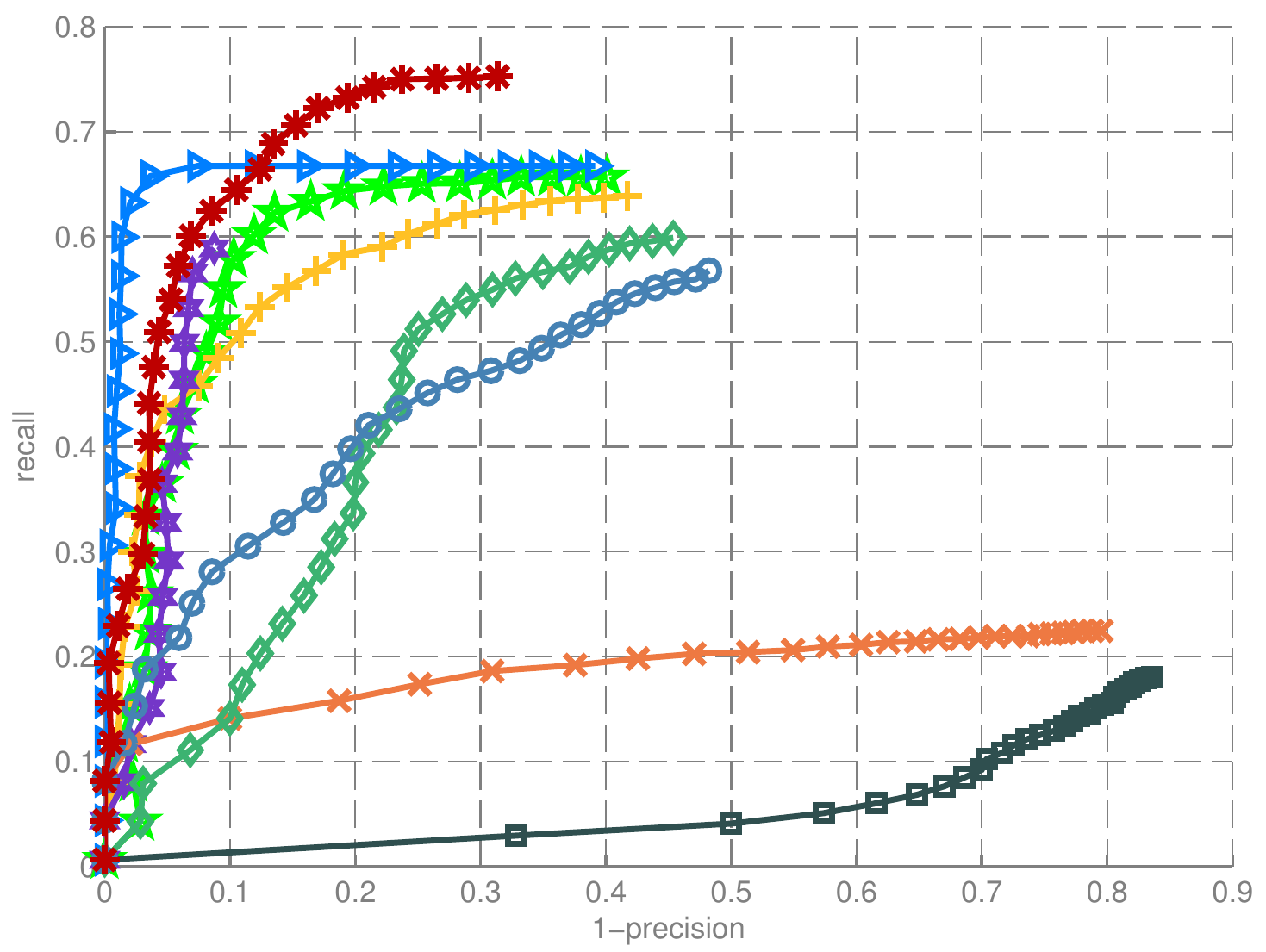} \vspace{-0.05in}\\%
(i) \tt{ miduomo01} & \hspace{-0.4 cm}(j) \tt{ townsquare} & \hspace{-0.4 cm}(k) \tt{ grafiti} & \hspace{-0.4 cm}(l) \tt{ tree} \vspace{0.05in}\\%
\end{tabular}
\includegraphics[width = 0.2 in]{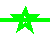} SM\ \ \ \includegraphics[width = 0.2 in]{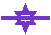} ACC\ \ \ \includegraphics[width = 0.2 in]{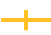} HV\ \ \ \includegraphics[width = 0.2 in]{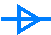} VFC\ \ \ \includegraphics[width = 0.2 in]{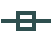} CAT\ \ \
		\includegraphics[width = 0.2 in]{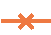} CAT+HV \includegraphics[width = 0.2 in]{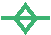} Ranking \includegraphics[width = 0.2 in]{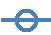} Ratio \includegraphics[width = 0.2 in]{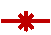} DE (Ours)
\end{center}
\caption{The precision-recall curves on $12$ image pairs. (a) $\thicksim$ (f) Six image pairs of the {\snu} dataset. (g) and (h) Two image pairs of the {\Rrwm} dataset. (i) and (j) Two image pairs of the {\sym} dataset. (k) and (l) Two image pairs of the {\vgg} dataset.}
\label{fig:ROC}
\end{figure*}

We firstly focus on the cases where a single descriptor is used. SIFT gives the best performance in dataset \snu, while LIOP performs best in dataset~\vgg. The experimental results show that the five descriptors complement each other and no single descriptor can get the best performance on all the four datasets. The optimal descriptor for matching vary from image to image. It hence points out that fusing multiple descriptors can be a feasible way for improving performance. As for the performances of the image matching algorithms, baseline HV averagely gets the superior results, since it fully supports multiple object matching, and stably works with various descriptors.

The four baselines for descriptor fusion, \ie CAT, CAT+HV, Ranking and Ratio, perform diversely. Baselines CAT and CAT+HV give poor performance. Even their accuracies in mAP fall behind those by the first four image matching algorithms that work with a single descriptor. It reveals that concatenation is not a good strategy for descriptor fusion, because worse descriptors degrade the discriminative power of the concatenated descriptor. Baselines Ranking and Ratio, especially Ratio, lead to much better matching results. The two baselines averagely outperform the four image matching algorithms, but still fall behind them if the best descriptor in each dataset is chosen. For instance, baseline Ratio gives $53.61\%$ in \snu~and $18.62\%$ in \Rrwm, while baseline ACC with SIFT achieves $60.28\%$ in \snu~and baseline HV with GB achieves $30.24\%$ in \Rrwm. In contrast, our approach allows mutual verification across different descriptors in an unsupervised manner, and correct correspondences will distinguish themselves with high coherence to each other. The quantitative results show that our approach can make the most of fusing various feature descriptors, and achieve significant performance gains over all the baselines on the four datasets.

\begin{figure*}[tH]
	\begin{center}
		\hspace*{-0.1 cm}	
			\begin{tabular}{ccc}
				\includegraphics[width = 2.2 in]{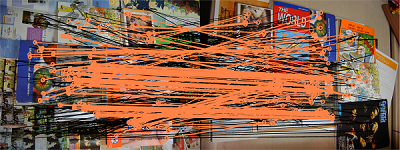}& \hspace*{-0.4 cm}	
				\includegraphics[width = 2.2 in]{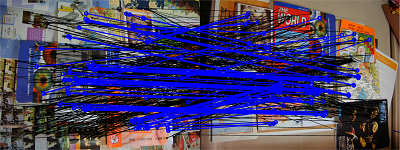} & 	\hspace*{-0.4 cm}	 \includegraphics[width = 2.2 in]{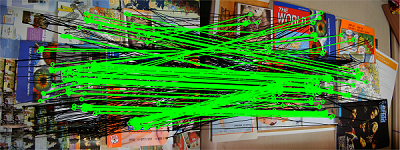}\\
				(a) SIFT (355/494) & \hspace*{-0.4 cm}	(b) LIOP (245/494) & \hspace*{-0.4 cm}	(c) DAISY (273/494)\\[5pt]
		
				\includegraphics[width = 2.2 in]{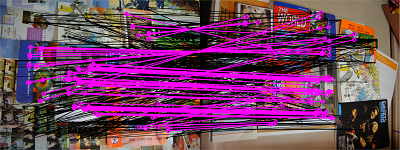} &\hspace*{-0.4 cm}	 \includegraphics[width = 2.2 in]{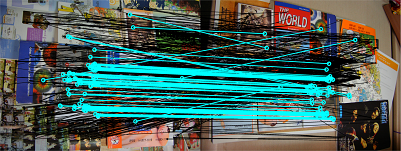}& \hspace*{-0.4 cm}	\includegraphics[width = 2.2 in]{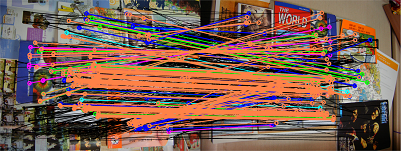}\\
				 (d) RI (175/494) & \hspace*{-0.4 cm}(e) GB (145/494) & \hspace*{-0.4 cm}(f) All (419/494)\\
		\end{tabular}	
	\end{center}
	\caption{The matching results by our approach, on image {\tt Books} of the {\snu} dataset, with (a) the SIFT descriptor, (b) the LIOP descriptor, (c) the DAISY descriptor, (d) the RI descriptor, (e) the GB descriptor, and (f) all the five descriptors. The recalls are shown in brackets. The correct correspondences are colored, and their colors indicate the descriptors by which they are established.}
	\label{fig:Descriptor2}
\end{figure*}

\begin{figure*}[tH]
	\begin{center}
	\hspace*{-0.1 cm}	
			\begin{tabular}{ccc}
				\includegraphics[width = 2.2 in, height = 0.8 in]{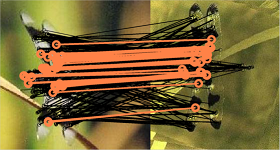}&\hspace*{-0.4 cm}	
				\includegraphics[width = 2.2 in, height = 0.8 in]{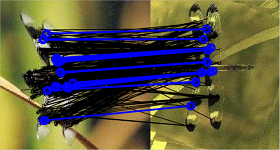} &\hspace*{-0.4 cm}	 \includegraphics[width = 2.2 in, height = 0.8 in]{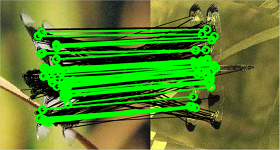}\\
				(a) SIFT (61/461) & \hspace*{-0.4 cm}	(b) LIOP (38/461) & \hspace*{-0.4 cm}	(c) DAISY (77/461)\\[5pt]
				
				\includegraphics[width = 2.2 in, height = 0.8 in]{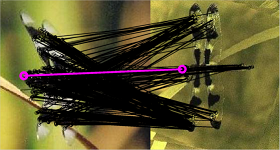} &\hspace*{-0.4 cm}	 \includegraphics[width = 2.2 in, height = 0.8 in]{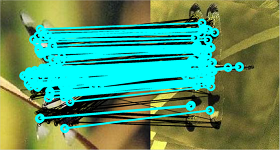} & \hspace*{-0.4 cm}	\includegraphics[width = 2.2 in, height = 0.8 in]{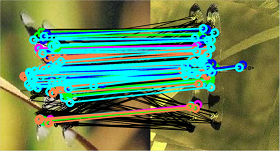}\\
				 (d) RI (3/461) & \hspace*{-0.4 cm}	(e) GB (132/461) & \hspace*{-0.4 cm}	(f) All (168/461)\\
		\end{tabular}	
	\end{center}
	\caption{The matching results by our approach, on image {\tt dragonfly} of the {\Rrwm} dataset, with (a) the SIFT descriptor, (b) the LIOP descriptor, (c) the DAISY descriptor, (d) the RI descriptor, (e) the GB descriptor, and (f) all the five descriptors. The recalls are shown in brackets. The correct correspondences are colored, and their colors indicate the descriptors by which they are established.}
	\label{fig:Descriptor1}
\end{figure*}

The mAP summarizes the performances of matching approaches on the whole dataset. To look inside how they work on individual images, precision-recall curves (precisely $1-$precision vs. recall curves here) are used. The {\snu} dataset consists of six image pairs. The resulting precision-recall curves by all the approaches are shown in~\figname~\ref{fig:ROC}(a) $\thicksim$ \ref{fig:ROC}(f). Note that we manually pick the best descriptor for baselines SM, ACC, HV and VFC to draw their curves for the sake of clearness. Thus, their performances may be overestimated in this sense. Baselines ACC and HV can deal with multiple object matching, while baseline VFC and SM are less robust in the cases. Ranking and ratio can increase the recall with the aid of multiple descriptors in most cases, but their precision is unsatisfactory. Our method can effectively match multiple objects, and considerably boost both the recall and precision by leveraging multiple descriptors.

We also select two pairs of images from each of the other three datasets, and plot the corresponding precision-recall curves in~\figname~\ref{fig:ROC}(g) $\thicksim$ \ref{fig:ROC}(l), respectively. The three datasets have different types of variations in matching, such as intra-class variations in dataset {\Rrwm}, combined changes in dataset {\sym}, and imaging condition changes in dataset {\vgg}. Our approach can deal with these variations by adaptively picking appropriate descriptors in matching interest points, and result in the superior performance. The exceptions are shown in \figname~\ref{fig:ROC}(k) {\tt grafiti} and \ref{fig:ROC}(l) {\tt tree}, two examples from dataset {\vgg}. As can be seen in \tabname~\ref{table:map}, descriptor LIOP achieves satisfactory results, and dominates the other descriptors on dataset {\vgg}. Hence, the performance gain of our approach is not significant in the cases.

\subsection{Visualization of Matching Results}

\begin{figure*}[tH]
\begin{center}
\hspace*{-0.1 cm}
\begin{tabular}{ccc}
\includegraphics[width = 2.2 in, height = 0.8 in]{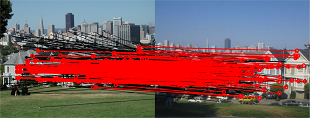} & \hspace*{-0.4 cm}
\includegraphics[width = 2.2 in, height = 0.8in]{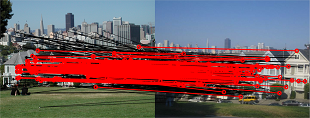} & \hspace*{-0.4 cm}
\includegraphics[width = 2.2 in, height = 0.8in]{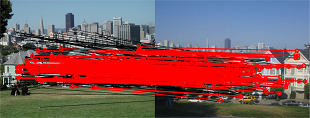}\\ %
(a) SM (323/661) & \hspace*{-0.4 cm}(b) ACC (306/661) & \hspace*{-0.4 cm}(c) DE (441/661)\\[5pt] %
\includegraphics[width = 2.2 in, height = 0.8in]{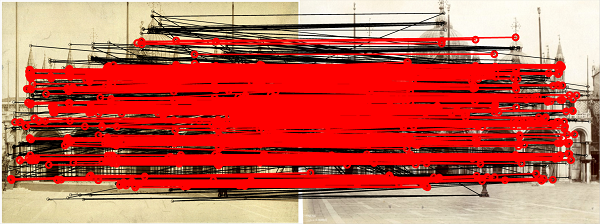} & \hspace*{-0.4 cm}
\includegraphics[width = 2.2 in, height = 0.8in]{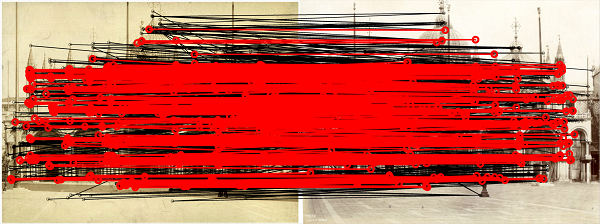} & \hspace*{-0.4 cm}
\includegraphics[width = 2.2 in, height = 0.8in]{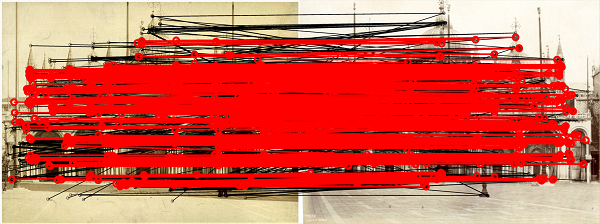} \\ %
(d) HV(652/1546) & \hspace*{-0.4 cm}(e) VFC (616/1546) & \hspace*{-0.4 cm}(f) DE (859/1546)
\end{tabular}
\end{center}
\vspace{-0.05in}
\caption{The matching results by our approach and the four image matching algorithms on two image pairs of the {\sym} dataset, including  (a) $\thicksim$ (c) image pair {\tt paintedladies12} and (d) $\thicksim$ (f) image pair {\tt trevi02}.}
\label{fig:Matching_SYM}
\end{figure*}

\begin{figure*}[tH]
\begin{center}
\hspace*{-0.1 cm}
\begin{tabular}{ccc}
\includegraphics[width = 2.2 in, height = 0.8in]{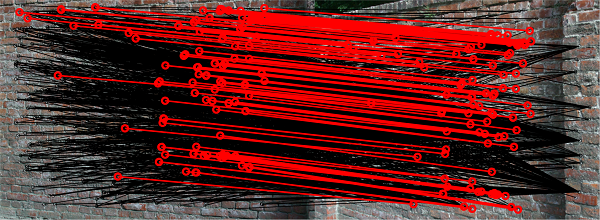} & \hspace*{-0.3 cm}%
\includegraphics[width = 2.2 in, height = 0.8in]{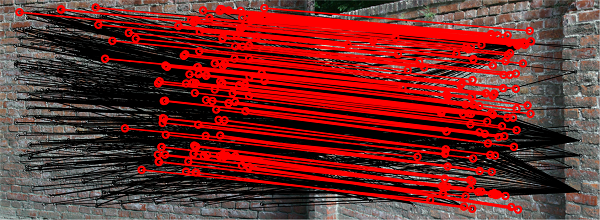} & \hspace*{-0.3 cm}%
\includegraphics[width = 2.2 in, height = 0.8in]{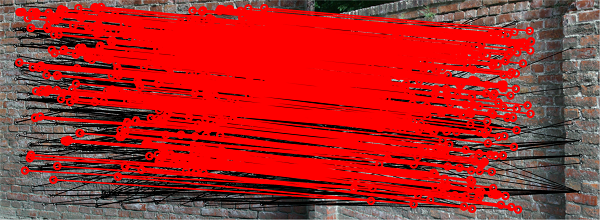}\\%
(a) CAT (160/1151) & \hspace*{-0.3 cm}(b) CAT+HV (198/1151) & \hspace*{-0.3 cm}(c)	DE (929/1151)\\[5pt]
\includegraphics[width = 2.2 in, height = 0.8in]{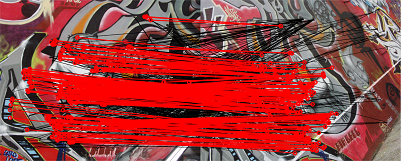} & \hspace*{-0.3 cm}%
\includegraphics[width = 2.2 in, height = 0.8in]{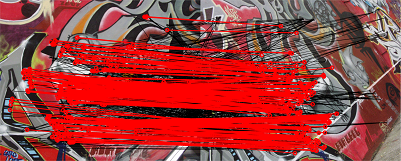} & \hspace*{-0.3 cm}%
\includegraphics[width = 2.2 in, height = 0.8in]{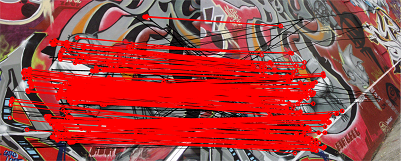}\\%
(d) Ranking (521/821) & \hspace*{-0.3 cm}(e) Ratio (537/821) & \hspace*{-0.3 cm}(f) DE (649/821)
\end{tabular}
\end{center}
\vspace{-0.05in}
\caption{The matching results by our approach and the four baselines for descriptor fusion on two image pairs of the {\vgg} dataset, including  (a) $\thicksim$ (c) image pair {\tt wall} and (d) $\thicksim$ (f) image pair {\tt grafiti}.}
\label{fig:Matching_VGG}
\end{figure*}

To gain insight into the quantitative results, we show the matching results by our approach and the adopted baselines on a few images. \figname~\ref{fig:Descriptor2} displays the matching results as well as the recalls by our approach, on image pair {\tt Books} of the {\snu} dataset, with each of the five adopted descriptors and all of them. The correct correspondences are colored with their colors indicating the descriptors by which they are established, \ie SIFT in orange, LIOP in blue, DAISY in green, RI in magenta, and GB in cyan. Descriptor SIFT finds the most correct matchings in this example. As shown in \figname~\ref{fig:Descriptor2}(f), our approach indeed selects most correspondences established by SIFT. Similar observation can be found in \figname~\ref{fig:Descriptor1}, in which the matching results on image pair {\tt dragonfly} of the {\Rrwm} dataset are plotted. In this example, descriptor GB performs best, and our approach also finds most correspondences by GB. The results in \figname~\ref{fig:Descriptor2} and \figname~\ref{fig:Descriptor1} show why our approach can leverage multiple, complementary descriptors to boost the matching performance: It adaptively determines the correct correspondences established by the better descriptors.

In \figname~\ref{fig:Matching_SYM}, the matching results by our approach (DE) and the four image matching algorithms on two image pairs, {\tt paintedladies12} and {\tt trevi02}, of the {\sym} dataset are shown. Our approach yields more dense and accurate matchings (red correspondences), and outperforms the four matching algorithms even if their respective best descriptors on this dataset have been manually chosen. In \figname~\ref{fig:Matching_VGG}, our approach and the four baselines for descriptor fusion are compared on two image pairs, {\tt wall} and {\tt grafiti}, of the {\vgg} dataset. Our approach in both cases carries out geometric verification, and effectively reduces the numbers of false positives (black correspondences) yielded by individual descriptors. Therefore, it achieves more satisfactory results.

To summarize, the visualization of the matching results demonstrates that our approach can effectively leverage multiple descriptors: On the one hand, it allows geometric layout verification across heterogeneous descriptors, and hence results in higher precision. On the other hand, it increases the number of correct correspondence candidates with the aid of complementary descriptors, and leads to higher recall. These properties enable our approach to alleviate the unfavorable issues in image matching, such as multiple objects with dramatic perspective variations in the {\snu} dataset, large intra-class variations in the {\Rrwm} dataset, the combined changes of lighting conditions and rendering styles in the {\sym} dataset, and imaging condition changes in the {\vgg} dataset.


\section{Conclusion\label{sec:con}}

We have presented an effective approach that can leverage multiple, complementary descriptors, and boost the performance of image feature matching. Specifically, the correspondences yielded by all descriptors are firstly projected into the homography space, in which both geometric and spatial consistency among the correspondences are measured by computing the geodesic distance on a designed graph. One-class SVM is then employed to rank the correspondences according to their consensus with each other. The proposed approach is featured with high flexibility in the sense that it can work with any elliptical region detectors as well as heterogeneous descriptors. Besides, it selects plausible correspondences across descriptors in a fully unsupervised way: no prior knowledge about images to be matched is required. Our approach has been comprehensively evaluated on four benchmark datasets, and compared with both the state-of-the-art approaches to feature matching as well as the baselines for descriptor fusion. The experimental results demonstrate that our approach can significantly boost the matching quality in both precision and recall, and is superior to the existing approaches and the baselines. In the future, we plan to generalize and apply this work to computer vision and image processing applications where accurate and dense matchings are appreciated, such as image alignment, object recognition, and motion estimation.


\bibliographystyle{spbasic}      
\bibliography{IJCVDE}   


%
%

\end{document}